\documentclass[letterpaper]{article} 
\usepackage{aaai25}  
\usepackage{times}  
\usepackage{helvet}  
\usepackage{courier}  
\usepackage[hyphens]{url}  
\usepackage{graphicx} 
\usepackage{natbib}  
\usepackage{caption} 
\usepackage{algorithm}
\usepackage{algorithmic}

\usepackage{newfloat}
\usepackage{listings}
\DeclareCaptionStyle{ruled}{labelfont=normalfont,labelsep=colon,strut=off} 
\floatstyle{ruled}
\newfloat{listing}{tb}{lst}{}
\floatname{listing}{Listing}
\title{Effort-aware Fairness: Incorporating a Philosophy-informed, Human-centered Notion of Effort into Algorithmic Fairness Metrics}
\author {
    Tin Trung Nguyen\textsuperscript{\rm 1}\thanks{Equal contribution.},
    Jiannan Xu\textsuperscript{\rm 2}\footnotemark[1],
    Zora Che\textsuperscript{\rm 1}, 
    Phuong-Anh Nguyen-Le\textsuperscript{\rm 3},
    Rushil Dandamudi\textsuperscript{\rm 1}, \\
    Donald Braman\textsuperscript{\rm 4},
    Furong Huang\textsuperscript{\rm 1},
    Hal Daumé III\textsuperscript{\rm 1},
    Zubin Jelveh\textsuperscript{\rm 3, \rm 5}
}
\affiliations {
    \textsuperscript{\rm 1}Department of Computer Science, University of Maryland, College Park, Maryland, USA\\
    \textsuperscript{\rm 2}Robert H. Smith School of Business, University of Maryland, College Park, Maryland, USA\\
    \textsuperscript{\rm 3}College of Information, University of Maryland, College Park, Maryland, USA\\
    \textsuperscript{\rm 4} George Washington University Law School, Washington DC, USA\\
    \textsuperscript{\rm 5}Department of Criminology and Criminal Justice, University of Maryland, College Park, Maryland, USA\\
    tintn@umd.edu, jiannan@umd.edu, 
    zche@umd.edu, nlpa@umd.edu, rushilcd@umd.edu, \\
    dbraman@law.gwu.edu, furongh@umd.edu, hal3@umd.edu, zjelveh@umd.edu
}

\usepackage{amsfonts}
\usepackage{amsmath}
\usepackage{graphicx}
\usepackage{subcaption}
\usepackage{booktabs} 
\usepackage{threeparttable} 
\usepackage{dsfont}
\usepackage{xcolor}
\usepackage{xspace}
\usepackage{amsmath}
\usepackage{soul}

\newcommand{\mycomment}[2]{\textcolor{#1}{\textbf{\textsf{\footnotesize[#2]}}}\xspace}
\newcommand{\tin}[1]{\mycomment{violet}{Tin: #1}}

\newcommand{\jiannan}[1]{\mycomment{blue}{Jiannan: #1}}

\newcommand{\hal}[1]{\mycomment{orange}{Hal: #1}}

\renewcommand{\hal}[1]{}
\renewcommand{\tin}[1]{}
\renewcommand{\jiannan}[1]{}
\begin{document}

\maketitle

\begin{abstract}
Although popularized AI fairness metrics, e.g., demographic parity, have uncovered bias in AI-assisted decision-making outcomes, they do not consider how much effort one has spent to get to where one is today in the input feature space. However, the notion of effort is important in how Philosophy and humans understand fairness.  We propose a philosophy-informed approach to conceptualize and evaluate Effort-aware Fairness (EaF), grounded in the concept of Force, which represents the temporal trajectory of predictive features coupled with inertia. Besides theoretical formulation, our empirical contributions include: (1) a pre-registered human subjects experiment, which shows that for both stages of the (individual) fairness evaluation process, people consider the temporal trajectory of a predictive feature more than its aggregate value; (2) pipelines to compute Effort-aware Individual/Group Fairness in the criminal justice and personal finance contexts. Our work may enable AI model auditors to uncover and potentially correct unfair decisions against individuals who have spent significant efforts to improve but are still stuck with systemic disadvantages outside their control.
\end{abstract}

%

\section{Introduction}
AI has assisted humans in making high-stakes decisions such as recidivism risk assessment \cite[e.g.,][]{bao2021s} and loan approval \cite[e.g.,][]{mayer2020unintended}, leading to potentially increased efficiency.\footnote{By 2023, $42\%$ of enterprises have actively deployed AI, while another $40\%$ are experimenting with the technology. Among those using or exploring AI, $59\%$ have accelerated their investments and implementation Efforts over the past two years \cite{ibm2025aidecision}.}
However, the growing deployment of AI in critical decision-making contexts has also raised serious concerns about algorithmic bias, which can disproportionately impact individuals and communities. In response, the AI literature has deeply explored two notions of fairness: group fairness (ensuring that a quantity of interest, such as accuracy, is equalized across demographic groups like race/sex) \cite[e.g.,][]{pedreshi2008discrimination} and individual fairness (ensuring that similar individuals get similar outcomes) \cite[e.g.,][]{dwork2012fairness}.

Typical applications of group and individual fairness treat similarly situated individuals as deserving similar treatment despite potential differences in demographic features (e.g., race, sex, and age). 
However, similarly situated individuals, e.g., those with the same income or arrest history, are often similarly situated for very different reasons. This leads to a hypothetical ``identical CVs'' problem: two people from very different backgrounds may have the same CV---and, therefore, are treated equivalently per group or individual fairness---but one of them may have had to work a lot harder to achieve that. The notion that Effort is an essential factor in fairness has been well studied in philosophy \cite[e.g., the free rider problem][]{hardin2003free} but has been largely, though not completely (see related work below), absent from the AI fairness literature. One key ingredient of Efforts is the temporal dimension: How an individual has changed over time. \citet{liu2018delayed} show that in an optimization problem, imposing a static fairness constraint, which often ignores the temporal dimension, may harm long-term improvement.

We first conceptualize a new measure of \emph{Effort-aware Fairness (EaF)}, grounded in philosophical literature \cite{de1803projet, massin2017towards, maine2002essai}, based on two key ideas: (1) acceleration, or more broadly the temporal trajectory of how an individual's task-relevant feature has changed over time; (2) inertia, or how much historical disadvantage each individual may have experienced outside their controls. 
We then conduct a human subjects experiment to validate whether our trajectory-based formulation of Effort is more relevant to laypeople's fairness perception than traditional formulations based solely on the feature's aggregate value \cite{romano2020achieving, ni2024fairness}. 
Finally, we demonstrate how to compute our EaF metric to audit AI models on two datasets, CLUE (criminal justice) and SHED (personal finance).\footnote{Our supplementary materials (e.g., data, code) are uploaded to a Google Drive folder: \url{https://drive.google.com/drive/folders/1RsSE1dUHDkn41iKCkIhY5QRdAVKVcrtB?usp=sharing}} 
\section{Background and Related Work}
\subsection{Traditional Metrics for AI Fairness}

In the technical CS/AI fairness literature, two main approaches mathematically formulate ``fairness'' based on the outcome of human/AI-assisted decision-making systems: group fairness and individual fairness. 
Group fairness, or group parity, is achieved when a statistical metric of interest (such as positive outcome rate or false positive rate) is equalized across different groups with respect to a sensitive feature, e.g., race or sex \cite{pedreshi2008discrimination, barocas-hardt-narayanan}.
Individual fairness, introduced formally by \citet{dwork2012fairness}, is based on the intuition that similar individuals should get similar outcomes. Formally, for a pair of individuals, given an input-space distance metric to measure how differently situated they are and an output-space distance metric to measure how different the distributions of their possible outcomes are, individual fairness requires (as a constraint in an optimization problem) that their output-space distance should be upper-bounded by their input-space distance. The more similarly situated two persons are (e.g., same demographics and criminal histories), the more likely that they should receive similar outcomes (e.g., risk scores). \footnote{We discuss the legal relevance of group fairness and individual fairness under U.S. law in Appendix A.} 

\subsection{Counterfactual ``Effort'' in AI Fairness Literature}
Recent AI fairness works attempt to incorporate some notion of ``Effort'' into fairness metrics by equalizing ``Effort'' across groups. They conceptualize ``Effort'' as the relationship between changes in mutable input features (e.g., prior arrests) and the probability predicted by a model of receiving a favorable outcome (e.g., granting bail). Approaches differ in terms of whether they take outcome parity as a requirement and attempt to equalize ``Effort'' across groups \citep{heidari2019long,guldogan2022equal} or whether they take equalized ``Effort'' as a requirement and attempt to equalize outcome parity \citep{gupta2019equalizing,huang2020fairness,von2022fairness}. These dual styles of approach are closely related to work on algorithmic recourse \citep{ustun2019actionable}, in that they ask how much effort'' an individual \emph{would have to put in} to achieve a favorable outcome. Both approaches can be loosely considered to be in the space of counterfactual ``Effort'' and in contrast to our approach---we are interested in how much ``Effort'' an individual \emph{has already put in} and use that as the value against which we seek parity.

Formally, existing approaches focus on counterfactual changes in input features ($\Delta \mathbf{x}$), which may lead to changes in the probability of receiving a favorable output label, e.g., from $f(\mathbf{x}) < \gamma$ to $f(\mathbf{x} + \Delta \mathbf{x}) \ge \gamma$ where $\gamma$ is a probability threshold. On the one hand, \citet{heidari2019long} and \citet{guldogan2022equal} set an upper threshold $\delta$ for the norm function $\mu$ of ``Efforts'', i.e. $\mu(\Delta \mathbf{ x}) \le \delta$, and seeks to equalize across groups the maximized favorable output rate. 
On the other hand, \citet{gupta2019equalizing}, \citet{huang2020fairness}, and \citet{von2022fairness} set a lower threshold for the desirable output rate and seek to equalize across groups the minimized norm of ``Efforts''. However, these works do not use actual changes over time, but they compare an individual with all others within the same group at only one time step to estimate $\Delta \mathbf{x}$ and $f(\mathbf{x} + \Delta \mathbf{x})$. \footnote{Shortcomings of the counterfactual Effort fairness literature and how our formulation overcomes them are in Appendix B.}

In summary, the recourse-based formulation of ``Effort'' in the CS/AI literature, though practically informative, has theoretical shortcomings and may not be rooted in more fundamental theories of Effort that philosophers have proposed, debated, and refined over centuries---a gap we seek to fill.

\subsection{How Philosophy Defines ``Effort''}
\label{subsec:philosophy}

As reviewed by \citet{massin2017towards}, the philosophy literature has used four main approaches to characterize Effort, of which two are directly related to Effort itself (rather than the perception of Effort):  (1) Force-based account (how much force would it take to ``move'' an object; introduced by \citet{de1803projet} and later formalized/advocated for by, i.a., \citet{maine2002essai}) and (2) Resource- or Energy-based account (spending Effort towards a goal means allocating part of a finite amount of resources, or energy tank, up until the energy tank is depleted; introduced by \citet{arai1912mental} and later formalized/advocated for by, i.a., \citet{gendolla2012Effort}). 

The Force-based account addresses several problems unresolved by the Energy-based account, such as capturing the concept of ``resistance'' to Effort, distinguishing between ``fatigue'' and ``Effort'', and explaining why Effort is unpleasant but praiseworthy. \citet{massin2017towards} informedly concludes that the Force-based account is ``more fundamental''.

The three main Philosophy arguments by \citet{massin2017towards} in support of the force-based account are highly applicable to our AI fairness contexts: (1) As ``only force-based accounts properly capture the idea of a resistance to our Effort,'' these accounts can capture issues such as systemic discrimination that resists individual-level progress \cite{bohren2022systemic}. (2) As ``force-based accounts, by contrast [to energy/resource-based accounts], stay clear of any substantive commitment about the relationship between fatigue and Effort,'' it is compatible with our AI-assisted decision-making contexts where a decision subject with better Effort (e.g., fewer arrests or higher income after each year) does not necessarily suffer from more fatigue. (3) As ``force-based accounts neatly explain the apparently paradoxical facts that we seem to praise and enjoy Efforts partly in virtue of their unpleasantness,'' it is compatible with the later introduced inertia and acceleration terms in our force-based formulation of Effort because having grown up in childhood poverty but still managing to work harder for better income (or desistance) might be unpleasant but indicating of virtue and thus praiseworthy \cite{mcneill2013moral}.
\citet{bermudez2023Efforts} later expand this force-based conceptualization of Effort to conceptualize the ``feeling of Effort'': ``Among the possible Effort-first accounts, a force-based approach seems most promising, partly because it allows for a direct explanation of why the feeling of Effort involves the experience of countering a resistance.''

Therefore, we will operationalize this philosophical force-based approach to formulate an individual-level ``Effort'' metric and incorporate it into existing algorithmic fairness metrics. 
Our force-based formulation of ``Effort'' will overcome both shortcomings of the recourse-based (counterfactual) formulation of ``Effort'' in the CS/AI literature: our approach is model-agnostic (as our force-based Effort metric is defined in a non-causal, output-independent sense) and we can loop force-based Effort into the individual distance function to define Effort-aware Individual Fairness (EaIF).
Furthermore, real-world fairness audits may necessitate quantification of past Efforts to inform current/future decisions. 
\section{Formulating Effort-aware Fairness (EaF)}
\label{sec:formulate}

To estimate ``Effort'' as an individual-level metric, we formulate the ``more fundamental'' Force-based account of ``Effort'' from philosophy \cite{massin2017towards} by using Newton's Second Law of Motion from Physics to develop an Effort-as-Force metric. Building on this idea, we formulate Effort-aware Fairness metrics to evaluate on real-world datasets.

\subsection{Effort}
\label{subsec:formulate_Effort}
As \citet{massin2017towards} does not give specific mathematical formulae to compute Force or Effort, we turn to the Physics literature for the well-known Newton's Second Law of Motion: $\mathbf{\mathbf{F}_{\text{net}}} = m \cdot \mathbf{a}$ where $\mathbf{F}_{\text{net}}$ is the net force applied on an object of inertia $m$ to give it an acceleration $\mathbf{a}$ (second-order time derivative of position $\mathbf{x}^i$) \cite{young2014university}.\footnote{$m$ is more commonly known as ``mass'' in everyday Physics. Inertia is a more generalizable understanding of this metric.}
We look for analogies to model the two quantities $m$ and $\mathbf{a}$ to compute our Effort-as-Force metric $E \propto \mathbf{F}_{\text{net}}$.
Assuming no friction, a person with inertia $m$ (characterized by systemic disadvantages or societal ``holding-back'' effects that are beyond their control) has to apply Effort $E$ to make their features (e.g., number of arrests or income) move in a desirable direction with an (observable) acceleration $\mathbf{a}$.

Our first quantity to model is inertia $m$. We conceptualize $m$ as the societal ``holding-back'' effect and/or historical disadvantage against an individual beyond their control/agency.
One example metric for inertia is hereditary disability, i.e., a disability passed down to a person by their parents. However, this feature might be hard to obtain (due to health privacy concerns) and difficult to quantify.
Another example metric for inertia is the socio-economic background of a person's family during their childhood, e.g., whether, back when the person was still a child, their family had to live below the federal poverty line, which is beyond the child's control. In our criminal justice and personal finance contexts, childhood poverty was found to be associated with significantly higher ``hazard rate of being convicted of violent criminality'' \cite{sariaslan2014childhood} and to have ``quantitatively large detrimental effects'' on adult earnings \cite{duncan2010early}. 

Next, acceleration $\mathbf{a}$ can be modelled as the second-order time derivative of task-relevant input features (e.g., number of past arrests in criminal justice, or income in personal finance). However, at first sight, there seems to be a problem with the second-order derivative. Suppose over a three-year period with equidistant time intervals ($\Delta t = \text{one year} = 1$), a person's annual income increases from $x_0 = 60k$ USD (first year) to $x_1 = 90k$ USD (second year) to $x_2 = 100k$ USD (third year). Our intuition suggests that they demonstrate positive Effort overall, as their annual income continued to increase, albeit at a slower velocity ($v_0 = 90k - 60k = 30k$ USD/year, followed by $v_1 = 10k$ USD/year). However, the slowing speed would result in a negative second-order derivative of $a_0 = 10k - 30k = -20k$ (USD/year$^2$), counterintuitively implying negative Efforts. 
Does this counterintuitive result mean the Force-based approach for Effort is wrong? Not necessarily. The underlying issue is that we should have modeled the temporal input feature (income) in a \textbf{cumulative} view:
$X_0 = 60k$ USD, $X_1 = 60k + 90k = 150k$ USD, and $X_2 = 150k + 100k = 250k$ USD.
In this cumulative formulation of income, velocity terms are $V_0 = 90k$ (USD/year) and $V_1 = 100k$ (USD/year); acceleration is $A_0 = 10k$ (USD/year$^2$), intuitively implying positive Efforts.
We use lowercase letters ($x,v,a$) to signal a non-cumulative view and uppercase letters ($X,V,A$) to signal a cumulative view.
This example highlights the importance of transforming a non-cumulative feature into a cumulative feature before computing its acceleration to quantify Effort. 

We justify formally our choice of a cumulative formulation of the Effort-related feature to compute acceleration. A non-cumulative version of a feature can be obtained by differentiating (taking the derivative of) the cumulative version of that same feature. Therefore, acceleration (second-order derivative, i.e., derivative of the first-order derivative) in the cumulative version corresponds to velocity (first-order derivative) in the non-cumulative version, e.g., $A_0=v_1$.
If we took the second-order derivative in the non-cumulative version, it would correspond to the third-order derivative in the cumulative version, which is theoretically at odds with force and hard to compute in practice.

Since Effort and AI-assisted decision making are social phenomena, we explain the social science relevance of our force-based formulation of Effort in Appendix C. 

\subsection{Effort-aware Individual Fairness (EaIF)}
To develop an Effort-aware Individual Fairness (EaIF) metric, we stipulate that similar individuals who spent similar Efforts should get similar outcomes. We propose a pairwise individual similarity function as a weighted combination between the aggregate task-relevant feature and Effort $E$.
We use Definition 2.1 (Lipschitz mapping) by \citet{dwork2012fairness} as the individual fairness condition. Let  $\mathbf{x}^i$ and $\mathbf{x}^j$ denote task-relevant feature vectors for a pair of individuals. A risk predictor $M$ maps individual $\mathbf{x}^i$ to a risk score, i.e., $M\mathbf{x}^i$. We define an Effort function $E(\cdot)$ that maps an individual $\mathbf{x}^i$ to an Effort score $E(\mathbf{x}^i)$ and an aggregate (summing) function $S(\cdot)$ (e.g., to sum all non-cumulative values of income or past arrests $S(\mathbf{x}^i)$ of individual $\mathbf{x}^i$). We also define an input-space individual distance function $d(\cdot,\cdot)$ over individuals' feature representations and an output-space distance function $D(\cdot,\cdot)$ over their predicted risk scores. The predictor $M$ achieves perfect Effort-aware individual fairness if for every pair of ($\mathbf{x}^i,\mathbf{x}^j$), this constraint is satisfied:
\begin{align}
D(M\mathbf{x}^i, M\mathbf{x}^j) \le d(\mathbf{x}^i,\mathbf{x}^j).
\label{eqn:IF}
\end{align}
Our EaIF formulation requires that the input-space individual distance function additionally takes into account the Efforts exerted by both individuals. With a scalar $0 \le \alpha \le 1$ to model the normative weight of Effort compared to aggregate features, we can give a simple formulation of $d$ based on the weighted Euclidean distance:
\begin{align}
d(\mathbf{x}^i,\mathbf{x}^j) = \sqrt{ \alpha [E(\mathbf{x}^i) - E(\mathbf{x}^j) ]^2 + (1-\alpha) [ S(\mathbf{x}^i) - S(\mathbf{x}^j) ]^2 }.
\label{eqn:pairwise_dist}
\end{align}
However, the AI community has recognized that individual fairness is hard to operationalize across different real-world contexts because it is hard to formalize an application-agnostic individual distance metric \citep[i.a.]{lahoti13operationalizing, ilvento2020metric}. We acknowledge that our formulation of $d(\mathbf{x}^i,\mathbf{x}^j)$ is just one among many options.

\subsection{Effort-aware Group Fairness (EaGF)}
\label{subsec:formulate_EaGF}
We formulate Effort-aware Group Fairness (EaGF) as $\hat{Y} \perp G \mid E$,  where $\hat{Y}$ is a predicted outcome (e.g., risk score) that predicts some ground-truth outcome $Y$ (e.g. violent recidivism, payment failure), $G$ is a protected demographic feature (e.g., race, sex, or age group, which defines the ``groups'' that we normatively expect parity), and $E$ is Effort as force. Our intuition behind this formulation is that people who have exerted similar amounts of Effort historically should receive similar predicted outcomes, regardless of their demographics. Computationally, EaGF means
\begin{align}
& \hat{Y} \perp G \mid E \nonumber \\
 \quad \Longleftrightarrow  \quad
& \forall e_0 \in \mathbb{R}, \; \forall g_i, g_j \in \mathcal{G},\nonumber \\
& \lim_{\epsilon \to 0} \mathbb{P}(\hat{Y} = 1 \mid E \in [e_0, e_0 + \epsilon], G = g_i) \nonumber \\
= & 
\lim_{\epsilon \to 0} \mathbb{P}(\hat{Y} = 1 \mid E \in [e_0, e_0 + \epsilon], G = g_j),
\label{eqn:EaGF}
\end{align}
where $[e_0, e_0 + \epsilon]$ is a small range (bin) of Effort values, which can be computed from inertia $m$ and acceleration $\mathbf{A}$ (from the cumulative formulation of Effort-related feature). The terms $g_i$ and $g_j$ are demographic groups (e.g., White and Black), and $\mathcal{G}$ denotes the set of all such groups.

\section{Datasets and Predictive Models}

For the following two empirical sections (human subjects experiment and metric demonstration pipelines), we use two real-world datasets that are relevant to two different AI-assisted decision-making contexts and have a temporal dimension. We also build simple predictive models to obtain AI predictions, on which Effort-aware Fairness can be evaluated.

\subsection{CLUE (criminal justice)}
\label{subsec:clue}
We select defendants from a real dataset, Client Legal Utility Engine (CLUE), 
with over $4.4$ million District Court and $700,000$ Circuit Court cases between $2012$ and $2020$ scraped by the Maryland Volunteer Lawyers Service (MVLS) from the Maryland Judiciary Case Search.\footnote{\url{https://casesearch.courts.state.md.us/casesearch/}} 
CLUE contains criminal history and demographic features, e.g., race, sex, and age (at the time of arrest). 

For our analysis, the key Effort-related feature is the number of prior arrests that resulted in charges before the current charge. The outcome variable is a risk binary label of getting re-arrested for a violent offense, within one year of the current arrest. Due to the data licensing condition, we cannot publicly release our clean version of CLUE. 

For every defendant who had an arrest between $2016$ and $2018$, we develop simple machine learning models (Random Forest classifier, whose predictions we use for the human subjects experiment, and three other models, LightGBM, Logistic Regression, and Decision Tree, whose predictions we use for the metric demonstration) with 5-fold cross-validation to predict their (violent) recidivism risk (on a scale of $0$ to $1$) within one year current their recent arrest.\footnote{We provide more details about the predictive models in Appendix D.}

\subsection{SHED (personal finance)}
We use the ``Survey of Household Economics and Decisionmaking'' (SHED) published by the Federal Reserve since 2013.\footnote{\url{www.federalreserve.gov/consumerscommunities/shed.htm}}
To obtain the time dimension, we perform record linkage across four years ($2019$ to $2022$). After selecting only the households that have data available across all four years, we obtain the final set of $704$ households. Our Effort-related feature is the annual range of household income (I40). We turn this range-based feature into a number by taking the lower bound. Our outcome feature of interest is the respondent's frequency of unpaid credit card balance (C4A), which can be of real-world predictive interest for financial institutions to process credit/loan applications.

We implement multiple AI models, including XGBoost, Random Forest, and Logistic Regression, to predict whether a household missed a payment on a credit card balance at least once in the $12$ months before the $2022$ survey. \footnote{Additional details about the predictive models are provided in Appendix D.}
\section{Human Subjects Experiment: Trajectory (Acceleration) Impacts Fairness Perception}
\label{sec:human_study}
\subsection{Experimental Design}

We must first decide on an appropriate population.
Several major works in the AI fairness literature have used laypeople's perception of fairness to decide on an appropriate fairness metric for a specific task, or even to construct new fairness metrics \cite{saxena2019fairness, harrison2020empirical, wang2019empirical}. 
We follow these precedents of informing our fairness metrics via laypeople's fairness perception, based on two rationales. 
In principle, laypeople's fairness judgment is vital for policy. \citet{awad2020crowdsourcing} argue that policymakers should be aware of and prepare for societal pushback of potentially beneficial but high-stakes decision-making AI tools. \citet{liu2021psychological} maintain that the discrepancy between a human-AI legal responsibility allocation framework from laws and the public's expectations may hinder the adoption of and trust in AI at large. 
In practice, to potentially achieve statistical significance, a human subjects experiment typically needs hundreds of participants. For example, the mean sample size for online human subjects experiments at CHI, a leading Human-Computer Interaction conference, was 224 participants \cite{caine2016local}. It is both expensive and logistically impractical to recruit such a large sample of technical CS/philosophy domain experts.

Next, we must decide on a more convenient fairness metric (e.g., group or individual fairness) to elicit fairness judgements from laypeople and measure the impact of Effort-related features on such judgements. We choose individual fairness over group fairness because individual fairness evaluation minimizes the effect of participants' cognitive load and mathematical ability as potential confounders.\footnote{If we validated the role of Effort in people's group fairness perception instead, we could not simply ask people to make fair decisions for two individuals, but for many individuals simultaneously from varying demographics to compute group parity metrics, increasing their cognitive load and making their fairness judgement more sensitive to non-substantive factors like visualization techniques \cite{van2021effect}.
A group fairness setting also assumes that people can compute group statistics (percentages) in their heads if they want to make group-wise fair decisions, which is not true even for college-level laypeople \cite{jacobs2018problem}.}

Individual fairness is formulated as "similar individuals are treated similarly" (Dwork et al., 2012). To evaluate whether a decision-making process (with or without AI) satisfies the individual fairness criterion, the probabilistic formulation of individual fairness might be simplified into two simple, more deterministic steps: (1) (Overall) Input-space distance: evaluate pairwise distance (based on features in the input space) to find pairs of highly "similar" individuals; (2) Output-space distance: in each pair of two highly similar individuals, check if their two outcomes are similar enough. 

\begin{figure}[htbp]
    \centering
    \includegraphics[width=0.45\textwidth]{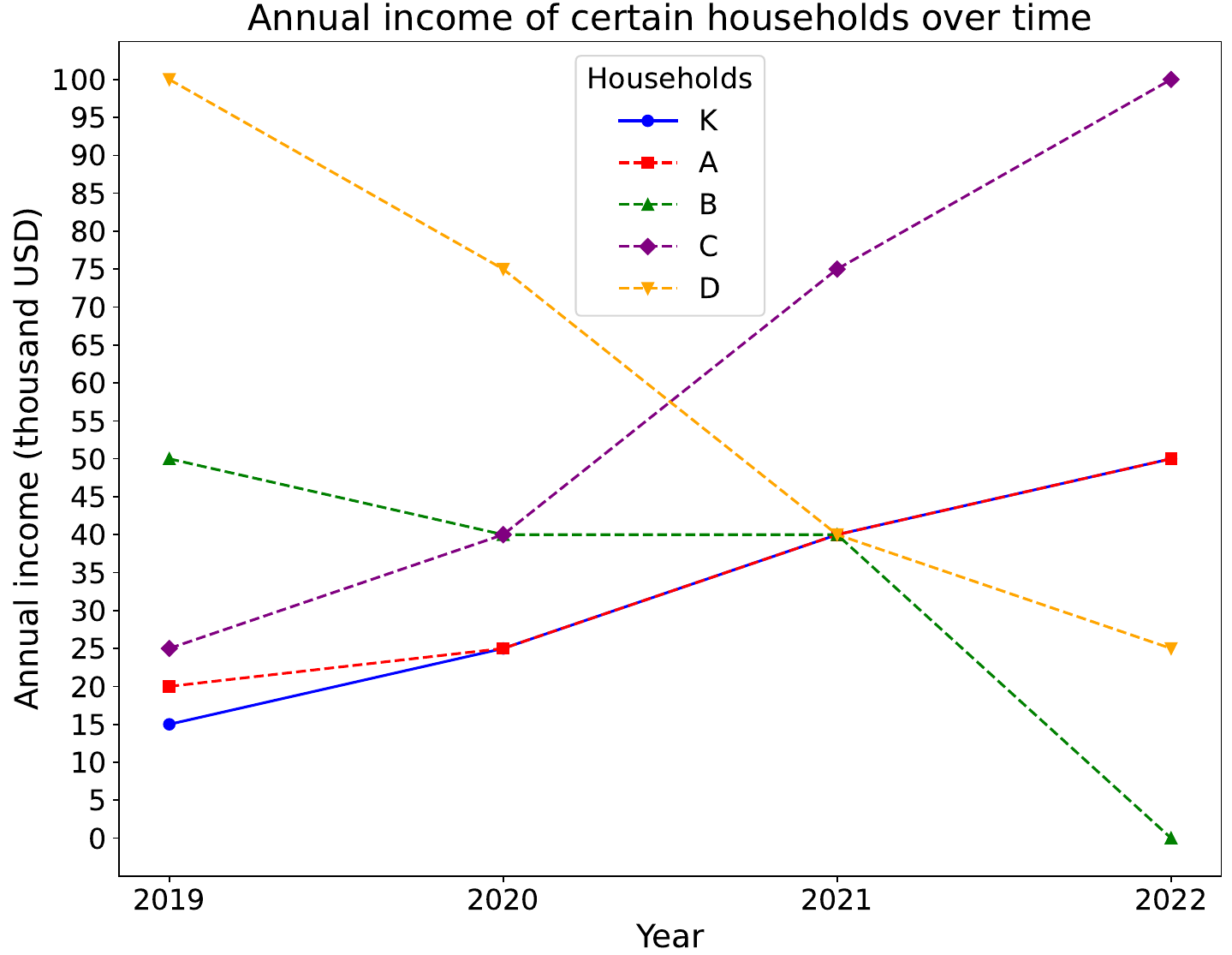}
    \caption{Example SHED set of 1 reference subject (K) and four subjects to be compared to K. A: similar trajectory/aggregate; B: different trajectory, similar aggregate; C: similar trajectory, different aggregate; D: different trajectory/aggregate. All households' incomes are real SHED records, except that we slightly perturb the first time step of K to get A.}
    \label{fig:SHED_trend1_KABCD}
\end{figure}

Our main question is whether people consider pairwise distance in terms of a feature's aggregate value ("aggregate distance") more or in terms of its trajectory ("trajectory distance") more during each of the two aforementioned steps in the individual fairness evaluation pipeline. 
We design two types of trajectory: increasing vs. decreasing, which means positive vs. negative velocity in a non-cumulative formulation (e.g., annual income) as shown in Figure \ref{fig:SHED_trend1_KABCD}, and is mathematically equivalent to positive vs. negative acceleration in a cumulative formulation (e.g., cumulative income since the first year). We show participants a non-cumulative (annual) feature view because many laypeople would not be able to distinguish positive vs. negative second-order derivative (convex vs. concave curvature) in a cumulative view.
Therefore, "trajectory" in this experiment corresponds to the cumulative-view "acceleration" term, a cornerstone in our philosophy-informed formulation of Effort. 
We draw real-world (anonymized) samples of defendants/households from CLUE (with past arrests as an Effort feature) and SHED (with income as an Effort feature). Our research questions:

\textbf{RQ1}: Does trajectory distance correlate with overall input-space distance? If so, does it correlate more strongly than aggregate distance? 

\textbf{RQ2}: Does trajectory distance correlate with output-space distance? If so, does it correlate more strongly than aggregate distance? 

\textbf{RQ3}: Do answers to RQ1 and RQ2 depend on the nature of the Effort feature, our between-subjects variable? (desirable ``income'' in SHED vs. undesirable ``arrests'' in CLUE)

The two main independent variables are trajectory distance and aggregate distance with respect to an Effort-related feature (e.g., number of arrests or income). We measure these variables by directly asking people to rate, on a Likert scale, the difference between two pairwise decision subjects in terms of their trajectories and in terms of their aggregate feature values (income/arrests summed across four years).

The two main dependent variables are overall input-space distance and output-space distance. 
Overall input-space distance is measured by asking participants, after they have rated the trajectory distance and aggregate distance for a pair, to rate, on the same Likert scale, how different the two decision subjects are overall.
Regarding output-space distance, in each set of $5$ decision subjects, $1$ of them is the reference subject ($K$) and the other $4$ are comparison subjects ($ A, B, C, D$).

The output-space distance for a given pair is then calculated as the absolute difference between the participant-assigned risk scores for the two subjects in the pair. For example, we calculate the output-space distance $D$ for the pair $(K, A)$ as:
\begin{align}
D(\tilde{M}\mathbf{x}^K, \tilde{M}\mathbf{x}^A) = \left| \tilde{M}\mathbf{x}^K - \tilde{M}\mathbf{x}^A \right|,
\end{align}
where $\tilde{M}$ presents participants' mental predictive model of the risk scores. 

To test if our findings will be robust across different AI application contexts and different Effort-related features, each participant will be randomly assigned to one of the two between-subjects conditions: CLUE (where the Effort-related, input-space feature is defendants' number of previous arrests, and the output-space feature is defendants' risk of committing a violent crime if released on bail) vs. SHED (where the Effort-related, input-space feature is households' income, and the output-space feature is households' risk of not paying back a monthly credit balance).

\subsection{Experimental Set-up}
\begin{figure*}[h]
    \centering
    \includegraphics[width=1.0\textwidth]{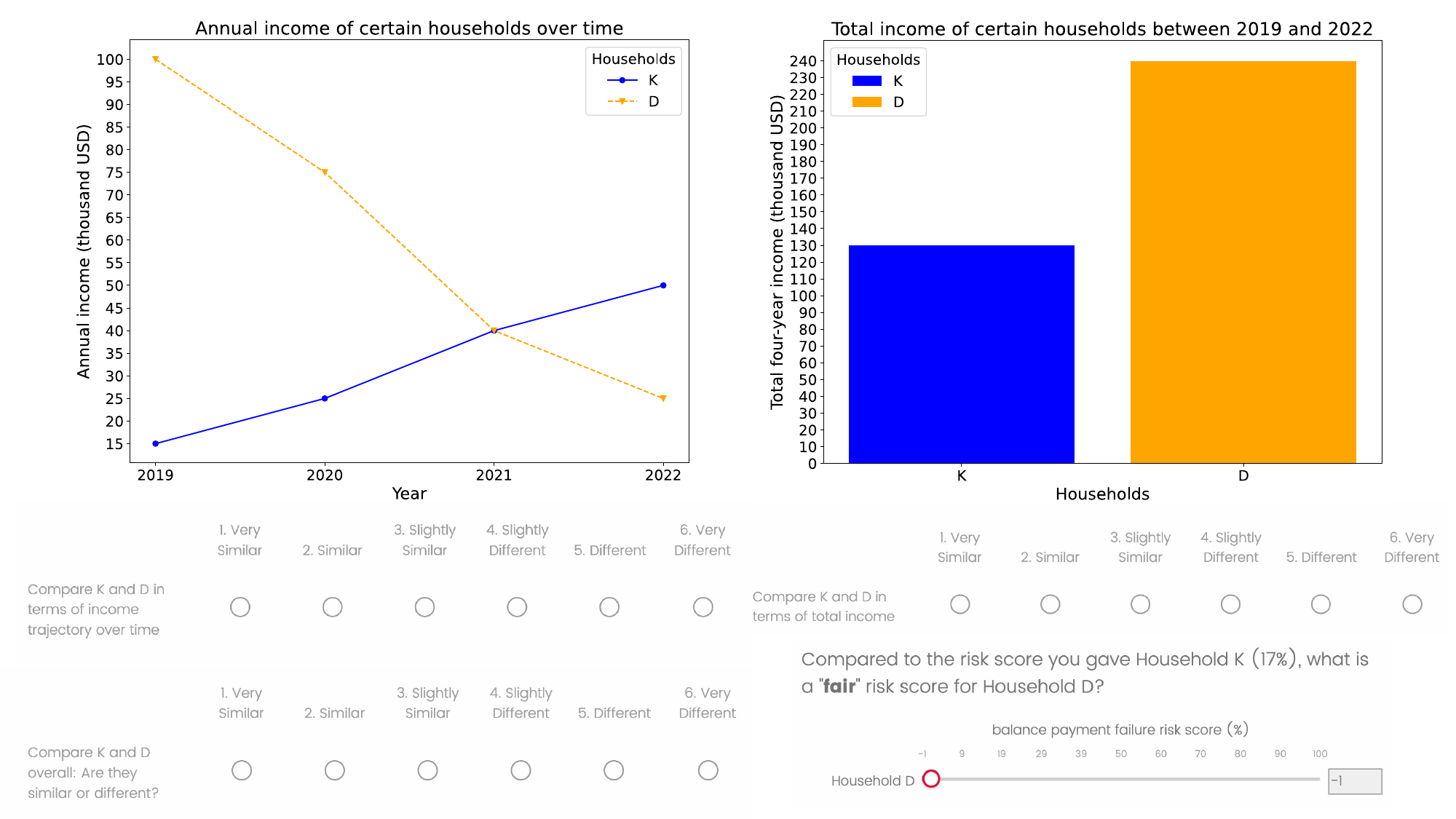}
    \caption{Example human subjects experiment interfaces to measure: (1) trajectory distance (upper left); (2) aggregate distance (upper right); (3) overall input-space distance (lower left); (4) output-space distance (lower right). \jiannan{All four panels correspond to the SHED condition and are shown to the same set of participants.}}
    \label{fig:SHED_interface}
\end{figure*}

We show interfaces for the main types of questions in the experiment in Figure \ref{fig:SHED_interface} (SHED condition) and Figure 4 (CLUE condition, Appendix E). Since the answers to these questions are highly subjective, to incentivize participants to think carefully and give informedly truthful opinions, we apply the Bayesian Truth Serum (BTS) method introduced by \citet{prelec2004bayesian} (details in Appendix E).
The BTS method was validated and applied in fields ranging from Psychology \cite{john2012measuring} and Marketing \cite{weaver2013creating} to AI \cite{witkowski2012robust} and HCI \cite{miller2014exploring}. Therefore, we use a simplified version of BTS on our most subjective type of MCQ: overall input-space distance, with a $\$5$ "Honesty bonus" for the top $10\%$ participants in BTS score. 

For the other type of dependent variable question (output-space distance), since this question asks people to estimate a numeric "fair" risk score, BTS (designed for MCQ) is not suited. Our alternative incentive mechanism is a $\$5$ "Rationale bonus" for the top $10\%$ participants who write the most detailed and persuasive rationales to justify their risk scores. As people write down justifications, they have more time to think and potentially recalibrate their risk scores.

We apply the following filters when recruiting Prolific participants: approval rate of at least $99\%$, Bachelor's degree, U.S. residency, and English fluency. Considering the median study completion time of approximately. $42$ minutes, we pay each participant $\$8.45$, corresponding to a median pay rate of $\$12$ per hour. After excluding $6$ participants who failed $2$ out of $3$ attention checks under Prolific's policy, our final data include $149$ samples.\footnote{144 regular participants and $5$ timed-out participants, who exceeded the maximum time allowed by Prolific but still finished the survey on Qualtrics}

This experiment has been approved by our Institutional Review Board (IRB).
We also pre-registered our experimental hypotheses and analyses on Jan $10$, $2025$  before recruiting Prolific subjects on Jan $11$, $2025$ to collect data.\footnote{Pre-registration proof: \url{https://aspredicted.org/4qg6-zpfm.pdf}}

\subsection{Analysis Methods}
To address RQ1, we compute Spearman correlation coefficients to examine the relationships between trajectory distance and overall input-space distance, as well as between aggregate distance and overall input-space distance. Spearman correlation is appropriate here because it does not assume linearity and is robust to non-normal distributions and outliers, which are suitable for ordinal data. We then perform one-sample t-tests to determine if these correlations are significantly different from $0$. If both correlations are significant, we use Hotelling's t-test \cite{may1997tests} and Steiger's Z-test \cite{meng1992comparing} to compare the strengths of these correlations. Hotelling's t-test is well-suited for assessing differences in dependent correlations, as it accounts for the shared dependent variable. However, it is known to be overly powerful, which can come at the expense of a higher Type I error rate in some scenarios \cite{may1997tests}. Steiger's Z-test, on the other hand, provides a more generalizable framework and is less sensitive to deviations from normality by implementing Fisher's z transformation of sample correlation coefficients. This makes Steiger's Z-test particularly valuable for larger datasets where the normality assumption may not hold perfectly. By using both tests, we can balance the strengths and limitations of each approach, providing more reliable comparisons of correlation coefficients. To address concerns related to multiple testing and reduce the likelihood of Type I errors, we apply the Bonferroni correction to adjust the p-values. To further validate these findings, we employ a bootstrapping procedure to estimate the confidence intervals of the differences between the two correlation coefficients. Moreover, we estimate a linear regression model:
\begin{align}
d & = \alpha_0 + \alpha_1 \cdot d_\text{S} +  \alpha_2 \cdot d_\text{E}  + \epsilon,
\end{align} 
where $d_{S}$ and $d_{E}$ represent the aggregate distance and the trajectory distance, respectively, and $d$ denotes the overall input-space distance. We conduct a Wald test with $H_0: \alpha_1 = \alpha_2  = 0$. If both $\alpha_1$ and $\alpha_2$ are significantly different from 0 and their effects are statistically significant, we can conclude that aggregate distance and trajectory distance have significantly different impacts on overall input distance. The rationale for including the linear regression model is that it allows us to examine the effects of aggregate distance and trajectory distance on overall input-space distance simultaneously. While Hotelling's t-test and Steiger's Z-test provide direct comparisons of the correlations, the linear regression offers additional insight into the magnitude and statistical significance of the individual contributions of aggregate and trajectory distances.  

For RQ2, we follow a similar approach as our aforementioned statistical tests in RQ1, simply replacing (pairwise) overall input-space distance $d$ with (pairwise) output space distance $D$. The corresponding regression model is:
\begin{align}
D & = \beta_0  + \beta_1 \cdot d_\text{S} +  \beta_2 \cdot d_\text{E}  + \epsilon,
\end{align} 
For RQ3, we perform a regression analysis to test if the between-subjects variable moderates the relationships examined in RQ1 and RQ2. We estimate the following models:
\begin{align}
d & = \alpha_0  + \alpha_1 \cdot d_\text{S} +  \alpha_2 \cdot d_\text{E}  +\alpha_3 \cdot C_\text{CLUE} \nonumber \\
& + \alpha_4 \cdot d_\text{S} \times C_\text{CLUE}  + \alpha_5 \cdot d_\text{E} \times C_\text{CLUE} + \epsilon,
\end{align} 
\begin{align}
D & = \beta_0  + \beta_1 \cdot d_\text{S} +  \beta_2 \cdot d_\text{E}  +\beta_3 \cdot C_\text{CLUE} \nonumber \\
& + \beta_4 \cdot d_\text{S} \times C_\text{CLUE}  + \beta_5 \cdot d_\text{E} \times C_\text{CLUE} + \epsilon,
\end{align} 
where $C_\text{CLUE}$ represents the between-subject variable,  indicating the experimental condition in which participants were randomized to the criminal justice context. Here, significant values for $\alpha_4$  and  $\alpha_5$  (or  $\beta_4$  and  $\beta_5$) indicate a moderation effect. These coefficients provide insights into whether respondents treat the relationships between aggregate or trajectory distances and overall input/output distances differently when evaluating criminal history versus income. For instance, if  $\alpha_4$  ($\beta_4$) is positive and significant, this suggests that respondents place greater weight on aggregate differences when evaluating criminal history compared to income. Similarly, if  $\alpha_5$  ($\beta_5$) is positive and significant, this indicates that respondents attribute more weight to trajectory differences when evaluating criminal history compared to income.

\subsection{Quantitative Results and Analysis}
Our responses to RQ1 and RQ2 are affirmative. In both the CLUE and SHED conditions, trajectory distance exhibits stronger correlations with overall input-space and output-space distances compared to aggregate distance, as shown in Table 2 (Appendix F). Under the CLUE condition, trajectory distance shows a strong correlation with overall input distance ($r_{\text{traj},\text{input}} = 0.7056$) and a moderate correlation with output distance ($r_{\text{traj},\text{output}} = 0.4218$). In contrast, aggregate distance has weaker correlations ($r_{\text{aggr},\text{input}} = 0.5022$ for input and $r_{\text{aggr},\text{output}} = 0.2089$ for output). In the SHED condition, trajectory distance strongly correlates with overall input distance ($r_{\text{traj},\text{input}}$ = 0.7881) and moderately with output distance ($r_{\text{traj},\text{output}}$ = 0.4986), while aggregate distance shows mixed correlations with a moderate correlation with input distance ($r_{\text{aggr},\text{input}} = 0.6547$) and a weak output correlation ($r_{\text{aggr},\text{output}} = 0.2507$). 

Table 3 (Appendix F) shows Hotelling's t-test and Steiger's Z-test to compare these correlations' strengths. Across conditions, trajectory distance consistently shows significantly stronger correlations with overall input and output distances than aggregate distance.\footnote{In the CLUE condition, the differences in correlations are highly significant, with Hotelling's t-test yielding statistics of $7.9758$ for input-space and $5.5161$ for output-space distances ($\text{p-value} < 0.001$). Steiger's Z-test further supports these results with statistics of $6.9077$ for input and $5.3461$ for output ($\text{p-value} < 0.001$). In the SHED condition, the differences are statistically robust, with Hotelling's t-test statistics of $7.3360$ (input) and $7.9612$ (output), and Steiger's Z-test statistics of $6.5541$ (input) and $7.6466$ (output), all significant at $\text{p-value} < 0.001$ after Bonferroni correction.} 
To further validate our findings, we implement a bootstrapping procedure in Table 4 (Appendix F) to estimate the confidence interval of the difference between the two correlation coefficients. \footnote{The $95\%$ confidence interval of the differences between $r_{\text{aggr},\text{input}}$ and $r_{\text{traj},\text{input}}$ is $[-0.2780, -0.1280]$ for CLUE and $[-0.1950, -0.0730]$ for SHED, with neither interval containing $0$. The $95\%$ confidence interval of the differences between  $r_{\text{traj},\text{output}}$ and $r_{\text{aggr},\text{output}}$ is $[-0.3000, -0.1280]$ for CLUE and $[-0.3330, -0.1610]$ for SHED. Both intervals also exclude $0$.}

As shown in Table 5 (Appendix F), after controlling for aggregate distance, trajectory distance remains significantly correlated with overall input-space distance\footnote{In CLUE, our regression model estimates $\alpha_1$ and $\alpha_2$ as $0.3580$ and $0.5454$, respectively, both significant at the $1\%$ level. The Wald test yields a chi-square statistic of $48.3438$ ($\text{p-value} < 0.001$). In SHED, $\alpha_1$ and $\alpha_2$  are estimated as $0.3182 $ and $0.6114$, respectively, both also significant at the 1\% level, with a Wald test chi-square statistic of $111.5533$ ($\text{p-value} < 0.001$).} 
and with output-space distance\footnote{In CLUE, our regression model estimates $\beta_1$ and $\beta_2$ as $0.0070$ and $0.0403$, respectively, both significant at the 1\% level, with a Wald test chi-square statistic of $39.4285$ ($\text{p-value} < 0.001$). In SHED, $\beta_1$ and $\beta_2$  are estimated as $-0.0132$ and $0.0565$, respectively, again both significant at the 1\% level, with a Wald test chi-square statistic of $116.2329$ ($\text{p-value} < 0.001$).}.

Regarding RQ3, while the CLUE vs. SHED condition itself does not have a direct impact on overall input distance ($\text{p-value} = 0.442$), it significantly moderates the impact of aggregate distance and trajectory distance. \footnote{The interaction terms $\alpha_4 = 0.0399$ ($\text{p-value} = 0.091$) and $\alpha_5 = -0.0661$  ($\text{p-value} < 0.001$) show that the effect of aggregate distance (and trajectory distance) on overall input-space distance becomes stronger (and weaker, respectively) in the CLUE (criminal history) condition than in the SHED (income) condition. A similar finding applies to output-space distance, considering $\beta_4 = 0.0201$ ($\text{p-value} < 0.001$) and $\beta_5 = -0.0163$ ($\text{p-value} < 0.001$).}

\subsection{Qualitative Results and Analysis}
After each input-space distance question, we have an optional question where participants select factor(s) that influence their assigned input-space distance: trajectory, aggregate, and ``Other (please specify)". Participants select trajectory more often than aggregate in both CLUE ($80.6\% > 71.4\%$) and SHED ($75.2\% > 60.0\%$) conditions, reaffirming our quantitative result that trajectory distance has a stronger impact than aggregate distance on overall input-space distance. We perform thematic coding inductively on the 55 free-text responses (rationales) that follow the third choice, ``Other (please specify below)" ($1.9\%$ in CLUE, $4.6\%$ in SHED), and come up with $9$ themes (after filtering out $4$ vague rationales). Although we expect people to specify ``Other" factors, most rationales are still about: (1) Both trajectory and aggregate ($23$ rationales); 
(2) Only trajectory ($12$ rationales); 
(3) Only aggregate ($3$ rationales). 

New themes that influence how people evaluate overall input-space distance include: 
(4) Concern about unobserved factors ($4$ rationales, e.g., ``Possible retirement or loss of job by B''); 
(5) First time step ($3$ rationales, e.g., ``Similar/near equal high amount of arrest to start with."); 
(6) Last time step ($2$ rationales, e.g., ``In addition to the vastly different trajectories, it is important to consider the endpoint of the graph, as it is at this time that the decision will be made, such as by a bank."); 
(7) Magnitude of acceleration ($2$ rationales, e.g., ``Jumps in increased or decreased crimes.''). 
Two other rationales are unique enough to each define its own theme: 
(8) Comparing to an average person: ``Although the total arrest times have a significant different, the number of total arrests for both people is quite a bit and more than an average person"; 
(9) Intersection: ``The intersection of their trajectories''. 

After each output-space distance question, participants may optionally explain their rationales, e.g., ``How did you decide on this `fair' risk score for A (compared to K)?". We manually code rationales for the first set (out of three defendant/household sets) to minimize the impacts of survey fatigue \cite{porter2004multiple}. After excluding vague, data-inconsistent, or AI-generated rationales, we have $243$ rationales (CLUE) and $194$ rationales (SHED). 

Among those rationales, our ``Only trajectory" code appears most often, assigned to $42.8\%$ of rationales in CLUE and $41.8\%$ in SHED (e.g., ``They both are trending upward which makes me think they are similar. I gave them both the risk of $28\%$ regardless of the number of arrests."). In contrast, our ``Only aggregate" code is assigned to merely $21.0\%$ of rationales in CLUE and $15.5\%$ in SHED (e.g., ``The graph shows that Household D has a significantly higher total income compared to Household K. Therefore, a fair risk score for Household D should be lower than $15\%$. [...]"). This qualitative comparison reaffirms our quantitative finding that trajectory distance influences output-space distance more strongly than aggregate distance does, with the caveat that many rationales still factor in both trajectory and aggregate distances ($21.4\%$ in CLUE and $30.4\%$ in SHED), e.g., ``Defendant D has been arrested a lot more than Defendant K, but he is trending down. That being said, I think it balances out and their likelihood of reoffending is probably similar."
\section{Demonstrating Effort-aware Fairness}
\label{sec:demo}
Our human subjects experiment shows that the key component in our force-based formulation of ``Effort'' -- the cumulative-view acceleration term (increasing vs. decreasing trajectories of non-cumulative, annual arrests/income) -- impacts laypeople's individual fairness evaluation process. We now demonstrate how to compute ``Effort'' and Effort-aware Individual Fairness (EaIF) on real-world datasets. 

Since the CLUE term of use does not allow us to release data, we will first compute our Effort, EaGF, and EaIF metrics on the published SHED dataset and release our record-linked version of SHED (2019-2022) for reproducibility. An analogous demonstration on CLUE data is in Appendix G. 

\subsection{Computing Effort}
One inertia metric that is relevant to both criminal justice and personal finance is childhood poverty, as explained earlier. However, for adults in our two real-world datasets, information on adults' childhood socio-economic background is not available, especially not at the individual level. Therefore, to quantify inertia empirically, we may use race as an individual-level but imperfect proxy for childhood poverty (or inertia) in the U.S. as the National Center for Education Statistics (NCES) found that childhood poverty rate differs substantially across races (e.g., White: $13\%$, Asian: $14\%$, Pacific Islander: $25\%$, American Indian: $36\%$, Black: $39\%$) in 2012.\footnote{\url{https://nces.ed.gov/programs/coe/indicator/cce/family-characteristics}—figure 4. Accessed Jan 19, 2025} To obtain $m$ per race, we scale these poverty rates by the maximum rate ($39\%$) so that $m$ spreads across $[0,1]$.

Given four equidistant time steps $\{0, 1, 2, 3\}$ (assuming $\Delta t = 1$), we model cumulative arrests in CLUE (or cumulative income, in multiples of $\$10k$ in SHED) at each time step as $X_0$, $X_1$, $X_2$, $X_3$. Their velocities are $V_0$, $V_1$, $V_2$, where $V_i = \frac{X_{i+1}-X_i}{\Delta t} = X_{i+1}-X_i$. Their accelerations are $A_0$ and $A_1$, where $A_{i} = \frac{V_{i+1}-V_i}{\Delta t} = V_{i+1}-V_i$. We average these two acceleration terms to get $A_\text{avg} = \frac{1}{2}(A_0 + A_1) = \frac{1}{2}(v_1 + v_2) $.
Noting that the Effort-related feature is desirable in SHED (income) but undesirable in CLUE (arrests), we compute the individual-level (good) Effort, $E$, with context adaptations:
\begin{align}
    \text{SHED: } & E = m \cdot \sigma(A_{\text{avg}}), \\
    \text{CLUE: } & E = m \cdot \big[1 - \sigma(A_{\text{avg}})\big].
\end{align}
The function $\sigma(z) = \frac{1}{1+e^{-z}}$ monotonically maps real values from $(-\infty, \infty)$ to the $(0,1)$ range, serving two purposes. First, it allows $\sigma(A_{\text{avg}})$ to obtain only one possible sign. More specifically, considering the desirable Effort feature, income, in SHED. Note that $m$ is always positive. When the average acceleration term is positive, $E = m \cdot A_{\text{avg}}$ will assign better (more positive) Effort to those with higher inertia (from more disadvantaged backgrounds). However, when the average acceleration term is negative,  $E = m  \cdot A_{\text{avg}}$ will assign worse (more negative) Effort to those with higher inertia, which contradicts our intention (given the same acceleration, more disadvantaged people should be credited for better Efforts). The sigmoid function eliminates the possibility of a negative sign and solves this problem. Second, the sigmoid function gathers extreme values closer together so that we have more data points to compute group parity even for subjects with extreme accelerations. 

\subsection{Computing Effort-aware Individual Fairness}
We operationalize the pairwise individual distance function (Equation \ref{eqn:pairwise_dist}) by first setting the weights, $\alpha$ and ($1-\alpha$), between the Effort feature and the aggregate feature to be either an equal baseline ($0.5$ and $0.5$) or our human subjects experiment's regression coefficients ($0.6114$ and $0.3182$ in SHED condition, which we normalize to $0.6577$ and $0.3423$ in this EaIF pipeline), illustrating the relative impacts by trajectory distance $d_E$ (proxy for difference in Efforts) and aggregate distance $d_S$ on the overall input-space distance $d$. 

To ensure the aggregate income feature stays in [0,1] like the Effort feature and reflect the intuition that the same income gap at lower income (e.g., $\$10k$ vs. $\$20k$) is more significant than at higher income (e.g., $\$110k$ vs. $\$120k$), we define an aggregate (sum) function $S(\mathbf{x})$ by applying a modified (scaled and translated) right-hand side of the sigmoid function $\sigma$ on the whole four-year (non-negative) income $X_3=\sum_{t=0}^{3} x_t$, scaled by $\lambda=\$200k$ so that this aggregate feature is empirically spread out across [0,1] range: 
\begin{align}
    S(\mathbf{x}) 
    = 2  \cdot \sigma\left(\frac{X_3}{\lambda}\right)-1.
\end{align}
Noting that the output-space and input-space distances ($D$ and $d$) take values in $[0,1]$, to operationalize the individual fairness constraint (Equation \ref{eqn:IF}), for each pair of individuals $\mathbf{x}^i$, $\mathbf{x}^j$ and predicted risks $M\mathbf{x}^i$, $M\mathbf{x}^j$ by an AI model $M$, we compute their pairwise fairness score:
\begin{align}
    F(\mathbf{x}^i, \mathbf{x}^j, M) = 1 - \max\{0, D( M\mathbf{x}^i, M\mathbf{x}^j ) - d( \mathbf{x}^i,\mathbf{x}^j ) \},
\end{align}
which should be 1 if this constraint is satisfied, or otherwise be lower, i.e., in $[0, 1)$, depending on the degree of constraint violation. We get an overall EaIF score for model $M$ by averaging pairwise fairness scores across all $\binom{\mathcal{|\mathcal{D}|}}{2}$ pairs from $|\mathcal{D}|$ individuals in dataset $\mathcal{D}$ (SHED) with predictions by $M$:
\begin{align}
    \operatorname{EaIF}(M) = \frac{1}{\binom{|\mathcal{D}|}{2}} \sum_{\mathbf{x_i}, \mathbf{x_j} \in \mathcal{D} \text{ and } i < j} F(\mathbf{x}^i, \mathbf{x}^j, M).
\end{align}
Table \ref{tab:EaIF_SHED} shows that according to our EaIF metric, XGBoost is less fair than the other two models, and random forest is slightly more fair than logistic regression. This result is consistent across both sets of weights (between the Effort feature and the most recent cumulative feature).

\begin{table}[htbp]
 \centering
 \caption{Effort-aware Individual Fairness Results on SHED}
 \label{tab:EaIF_SHED}
\begin{tabular}{lcc}
\toprule
           & \multicolumn{2}{c}{\textbf{EaIF}} \\
\textbf{Model}                  & Equal weights    & Human study weights  \\
\midrule
XGBoost             & 0.80                             & 0.79                                     \\ 
Logistic regression & 0.88                            & 0.87                                     \\ 
Random forest      & 0.91                            & 0.90 \\ 
\bottomrule
\end{tabular}
\end{table}

\subsection{Computing Effort-aware Group Fairness}
\label{section:EaGF}

\citet{binns2020apparent} found that group fairness and individual fairness are rooted in similar philosophical concepts (``consistency'' and ``egalitarianism''). Algorithms that satisfy both group fairness and individual fairness were developed, demonstrating the two metrics' compatibility \cite{zemel2013learning}. Therefore, we will also demonstrate how to compute Effort-aware Group Fairness (EaGF) with our same force-based notion of ``Effort'', which was validated by our human subjects experiment in the individual fairness context.

Given an individual-level Effort feature, our goal is to compute Effort-aware group parity and observe this metric as a function of numeric Effort.
Our main steps include (1) partitioning decision subjects into similar-effort bins; (2) computing a demographic group parity metric within each bin; (3) plotting this parity metric against the Effort bins. We used demographic features of each household representative who completed the SHED survey to compute group fairness metrics.

After computing a (good) Effort value for every individual, we partition all individuals and their Effort values into equidistant bins (intervals) of Effort length = 0.1. Within each bin, which represents individuals with similar Efforts, we compute the demographic group parity in terms of the mean predicted risk by the same model for each demographic group ($G$ is race, sex, or age).
For statistical stability, we only consider a demographic group if the group has at least 10 data points in the respective Effort bin. We define (within-bin) group parity (scale: 0-1) as the ratio of the demographic group with the lowest average risk score $\hat{Y}$ divided by the demographic group with the highest average risk score $\hat{Y}$, formally derived from Equation \ref{eqn:EaGF} as follows:
\begin{align}
\label{eqn:EaGF}
\text{Parity}_{G, e_0} = \frac{\min_i \mathds{P}(\hat{Y} = 1 \mid E \in [e_0, e_0 + \epsilon] ,  \: G = g_i)}{\max_j \mathds{P}(\hat{Y} = 1 \mid E \in [e_0, e_0 + \epsilon] ,  \: G = g_j)}.
\end{align}

\begin{figure}[htbp]
    \centering
    \begin{subfigure}[b]{0.45\textwidth}
        \centering
        \includegraphics[width=\textwidth]{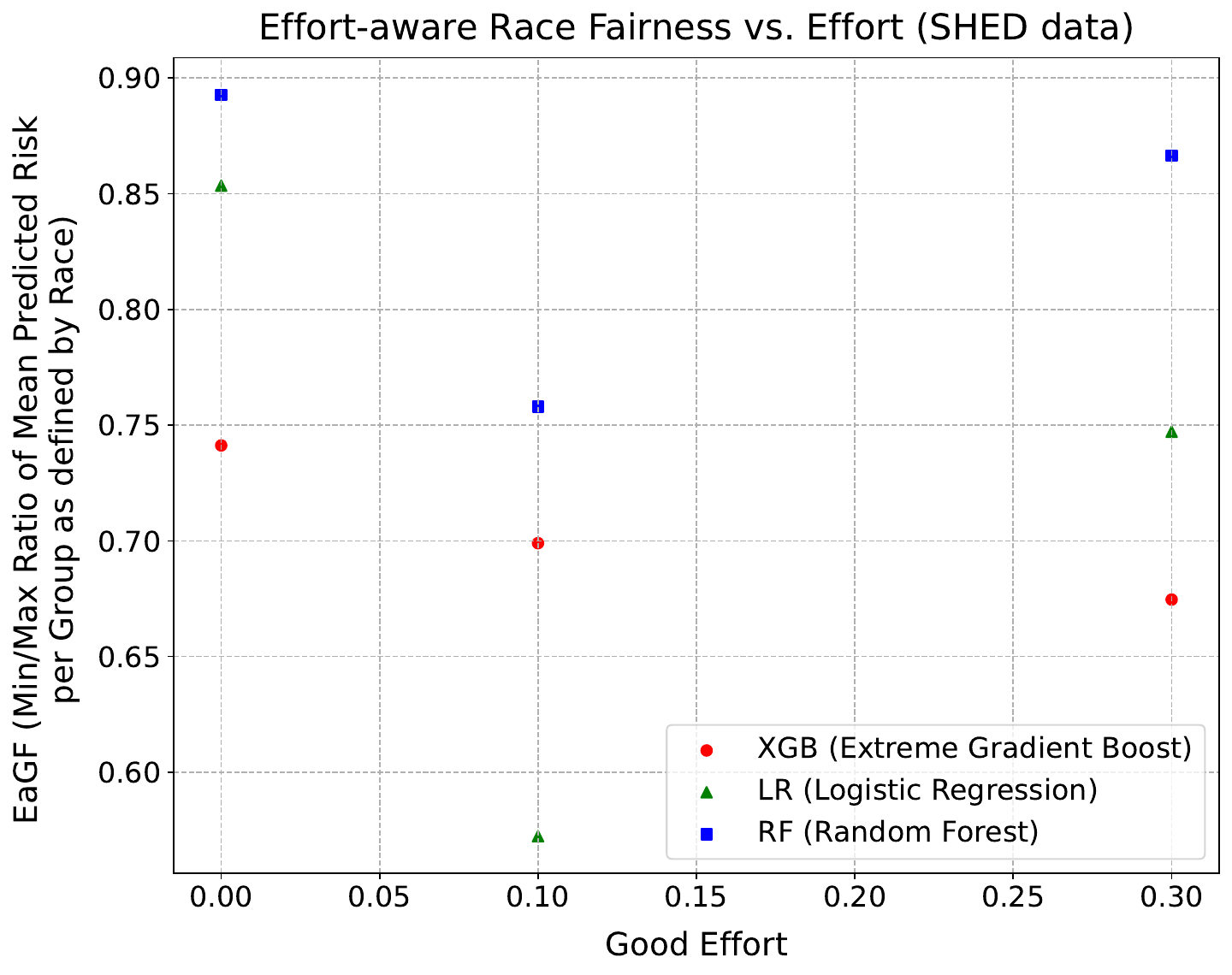}
        \caption{Race parity}
        \label{subfig:SHED_EaGF_App1_Race}
    \end{subfigure}
    \hfill
    \begin{subfigure}[b]{0.45\textwidth}
        \centering
        \includegraphics[width=\textwidth]{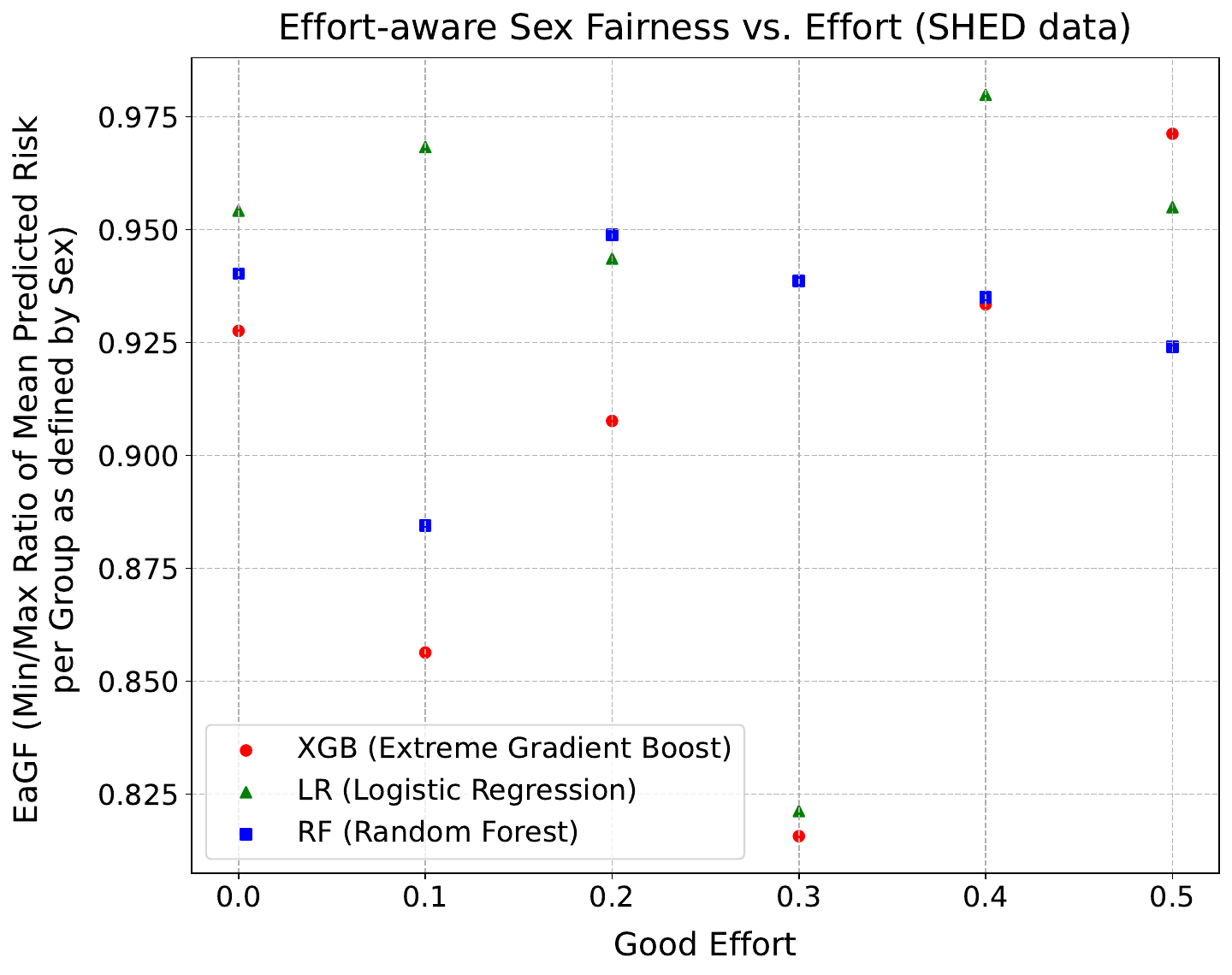}
        \caption{Sex parity}
        \label{subfig:SHED_EaGF_App1_Sex}
    \end{subfigure}
    \hfill
    \begin{subfigure}[b]{0.45\textwidth}
        \centering
        \includegraphics[width=\textwidth]{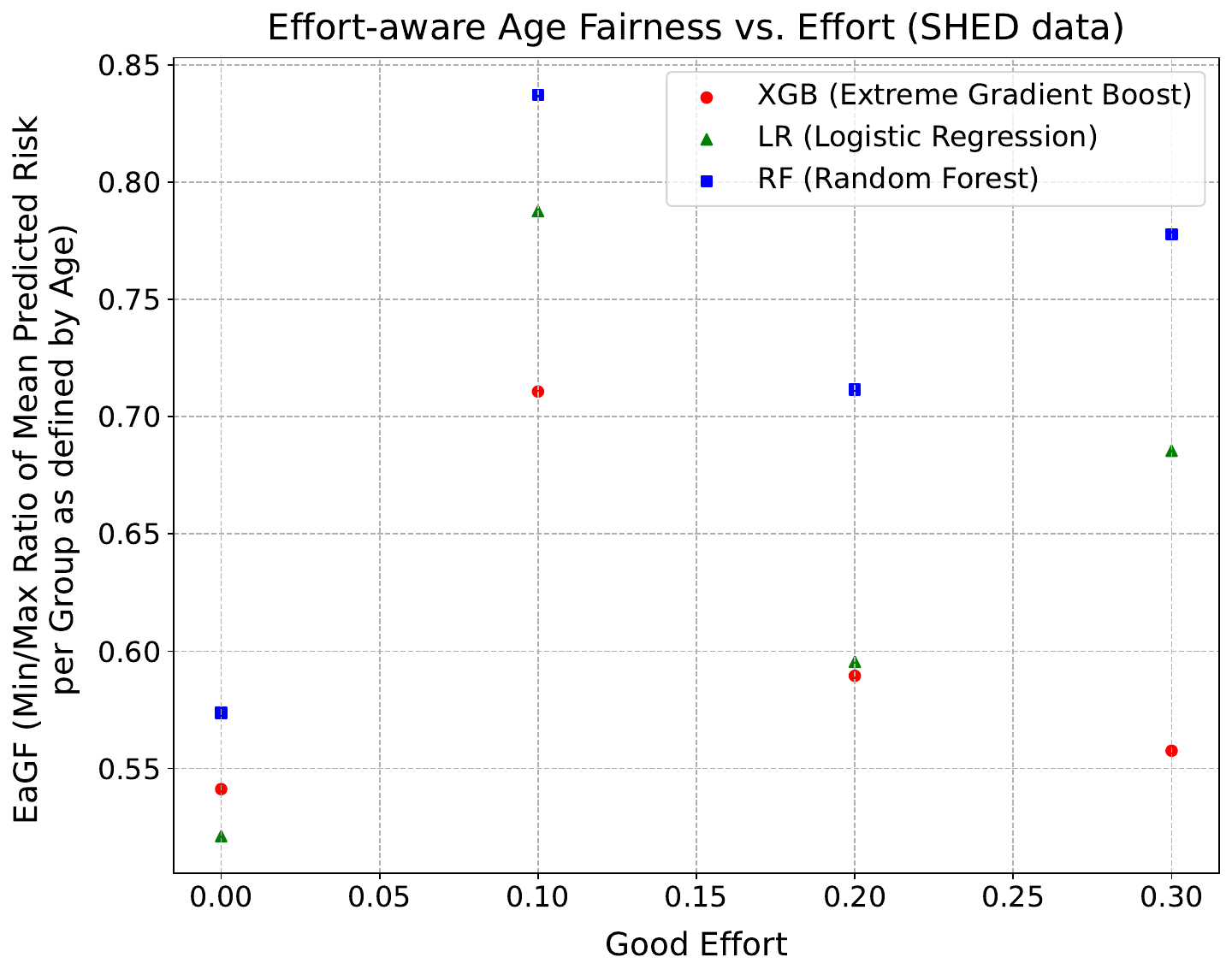}
        \caption{Age group parity}
        \label{subfig:SHED_EaGF_App1_Age}
    \end{subfigure}
    \caption{Effort-aware Group Parity (in terms of mean predicted risk) as a function of Effort (SHED data). Bin length is 0.1. Only demographic groups with at least 10 data points are considered. The subplots' x-axes might differ since not all Effort bins within the 0-1 range have enough data points to compute EaGF.
    Random forest consistently achieves the best EaGF results in terms of race and age, but not sex.}
    \label{fig:SHED_EaGF_App1}
\end{figure}
Figure \ref{fig:SHED_EaGF_App1} shows Effort-aware, i.e., within-bin, demographic group parity as a function of Effort. To illustrate why EaGF information is helpful, we first look at Table 6 (Appendix F) on overall demographic parity, i.e., not conditioned on similar Efforts, showing random forest to be among the most group-wise fair models in terms of race, sex, or age group. Figures \ref{subfig:SHED_EaGF_App1_Race} and \ref{subfig:SHED_EaGF_App1_Age} reassure us that, conditioned on similar Efforts, random forest remains the fairest model among racial and age groups. However, Figure \ref{subfig:SHED_EaGF_App1_Sex} cautions us that when conditioned on higher-effort bins ($0.4$-$0.5$ and $0.5$-$0.6$), random forest is among the least fair models in terms of sex. This pattern might be concerning since we do not want to disincentivize people from exerting more Effort. If sex parity is a legal/policy priority in our context, our EaGF audit might suggest using XGBoost instead of random forest because XGBoost achieves roughly the same best overall sex parity as random forest (Table 6) and XGBoost achieves the best EaGF score when conditioned on the highest-Effort bin (Figure \ref{subfig:SHED_EaGF_App1_Sex}), thereby incentivizing people to exert more effort. Our demonstration highlights a key strength of EaGF over traditional group fairness, as it provides a more granular view of any concerning trend in parity vs. Effort when evaluating AI models.

\section{Conclusion}
We identify a gap between how the CS/AI literature has attempted to formalize the notion of ``Effort'' into existing fairness metrics and how philosophers/legal scholars understand ``Efforts''. We seek to fill this gap by operationalizing the Force-based account of Effort from philosophy, formulating an ``Effort'' metric as the product of inertia (systemic disadvantage beyond an individual's control, such as childhood poverty) and accelerations (temporal changes in a task-related input-space feature), which can be looped into existing fairness metrics to formulate Effort-aware Group/Individual Fairness (EaGF/EaIF). As we can conceptualize acceleration more broadly as trajectory, we conduct a human subjects experiment and successfully validate that the acceleration/trajectory component of our Effort-aware Fairness formulation aligns with laypeople's perception of fairness across two distinct AI-assisted decision-making contexts, namely criminal justice and personal finance. Finally, we develop practical pipelines to compute EaIF and EaGF on real-world datasets (SHED and CLUE), demonstrating how these metrics complement traditional fairness metrics.  
\section{Ethics Statement}
We minimize privacy risks to our human subjects experiment participants by removing any identifiable information (such as Prolific ID) from any supplementary materials we publish, as well as complying with any related institutional (IRB) and federal human subjects protection policies. We commit to compensating our participants reasonably. For example, as our original estimation of 30 minutes for study completion time turns out to be an under-estimation compared to the median completion time (42 minutes), we have adjusted the pay rate from $\$6$ to $\$8.45$ to maintain a median hourly compensation rate of $\$12$, which is considered a ``Good'' rate in Prolific standard. 
We also carefully comply with the CLUE dataset's terms of use to protect defendants' privacy, such as by storing the dataset on a password-secured server and only using non-personally identifiable features in our simple AI models development and fairness evaluation.

We acknowledge two core limitations of our proposed metrics. 
First, in our formulation of both EaIF and EaGF, by using efforts to find pairs of similar individuals or by conditioning on similar-effort bins when computing group parity metrics, we normatively prescribe the ideal outcomes of similar-effort individuals. Still, we leave open the more complex normative question of how the outcomes of individuals with different levels of Effort should be. We have not incorporated a notion of directionality for the relationship between Effort and outcome. In other words, we do not necessarily stipulate that people who exert more Effort should always receive more favorable outcomes, as this involves more unsettled philosophical debates on meritocracy, e.g., whether resources should be allocated based on one's efforts or one's natural talents (such as a basketball athlete's height), which are often due to luck and beyond one's control \cite{sandel2021meritocracy}.
Second, the risk of fairwashing, i.e., generating posthoc explanations to make predictions by an otherwise unfair black-box AI models score well on specific fairness metrics \cite{aivodji2019fairwashing, aivodji2021characterizing}, might extend to our EaF metrics because bad actors might manipulate the particular time steps when computing our acceleration term as well as the feature(s) from which our inertia term or the societal holding back effect is approximated. Therefore, we welcome future research on how to guard the quantification of Efforts against fairwashing, as well as more AI-assisted decision-making data collection initiatives to include more individual-level features that might reflect disadvantages beyond one's control, such as disability or childhood poverty.

\section{Acknowledgments}
This material is based on work supported in part by the Institute for Trustworthy AI in Law and Society (TRAILS), which is supported by the National Science Foundation under Award No. 2229885. Any opinions, findings, and conclusions or recommendations expressed in this material are those of the author(s) and do not necessarily reflect the views of the National Science Foundation or the National Institute of Standards and Technology.
This work is also supported in part by the National Institute of Justice's Graduate Research Fellowship (Award No. 15PNIJ-23-GG-01932-RESS).

\bibliography{aaai25}

\appendix
\clearpage
\onecolumn

\section{Appendix A: Judicial Interpretations of Equal Protection: Bridging Group and Individual Fairness}
Case law deems both group fairness and individual fairness as relevant to the constitutional ``Equal Protection'' clause. For example, the U.S. Supreme Court (SCOTUS) ruled in Washington v. Davis (1976) that: ``Disproportionate impact is not irrelevant, but it is not the sole touchstone of an invidious racial discrimination forbidden by the Constitution.'' \cite{1976washington}. Older SCOTUS decisions such as F.S. Royster Guano Co. v. Commonwealth of Virginia (1920) interpret ``Equal protection'' consistently with individual fairness: ``The Equal Protection Clause of the Fourteenth Amendment commands [...] that all persons similarly situated should be treated alike'' \cite{1920fs}. Interestingly, a synergy of both group and individual fairness is found in City of Cleburne, Tex. v. Cleburne Living Center (1985): ``Discrimination, in the Fourteenth Amendment sense, connotes a substantive constitutional judgment that two individuals or groups are entitled to be treated equally with respect to something'' \cite{1985cleburne}. 

\section{Appendix B: Discussion about Counterfactual Effort in AI Fairness Literature}
Our primary concern with this AI Fairness literature is that ``Effort'' was framed but should not have been framed in a counterfactual sense as recourse or the distance (in the input feature space) between a decision subject's current position and the decision boundary (which separates feature vectors that get good outcomes from feature vectors that get bad outcomes). We recognize the importance of these counterfactual ``Effort'' fairness or recourse fairness metrics, such as ``equal improvability''  \cite{guldogan2022equal}, which often seek to equalize the distance to the decision boundary across demographic groups. 

However, defining Effort in terms of recourse has two theoretical shortcomings. First, recourse might vary depending on the AI model, as the decision boundary will change with the model. However, Effort should be model-agnostic and should depend only on the personal history of each individual. Second, ``Effort'' as recourse cannot be incorporated into the individual fairness regime established by \citet{dwork2012fairness} because recourse is only well-defined for decision subjects receiving bad outcomes, not for decision subjects already receiving good outcomes.

Suppose we incorporate ``Effort'' as recourse into the individual distance function (Step 1 of individual fairness evaluation). If that distance function identifies two individuals with highly similar recourse, those recourse values must be well-defined and the outcomes of those two individuals must be the same (both good), automatically guaranteeing a perfect individual fairness score. We give an example to illustrate the incompatibility between recourse and the individual fairness regime: \citet{von2022fairness} defines an ``Individually Fair Causal Recourse'' metric as ``if the cost of recourse would have been the same had the individual belonged to a different protected group.'' However, there is no individual distance function defined, and this metric is more rooted in procedural fairness (a protected demographic feature should not impact a quantity of interest, such as cost of recourse) than in ``individual fairness'' as defined by \citet{dwork2012fairness}. \jiannan{Why is it rooted in procedural fairness but not group fairness? The way we explain here is more like a group fairness criterion but not procedural fairness...}

\section{Appendix C: Social Science Grounding of Effort-aware Fairness }
\citet{inzlicht2018Effort} (cognitive science) define ``Effort'' in their glossary as ``intensification of either mental or physical activity in the service of meeting some goal (e.g., increasing the \textit{force} applied to an object).''
Several works in the psychology literature also measure Effort as force: ``Kruglanski et al (2012) are also quite clear that Effort (effective driving force) equals the demands (resistive force) but cannot exceed capacity (potential driving force)'' \cite{steele2020perception, kruglanski2012energetics}.
In the criminology literature specifically, both components in our force-based formulation of Effort, acceleration (which represents temporal trajectory more broadly) and inertia, have been extensively studied. Our acceleration (trajectory) term is grounded in the ``criminal career paradigm'' formalized by \citet{piquero2003criminal}, which studies the prevalence, frequency, and desistance of criminal offenders over time to predict recidivism. In fact, the concept of ``acceleration'' was defined in this paradigm: ``acceleration, which refers to an increase in the frequency of offending over time'' \cite{piquero2003criminal}. Furthermore, ``regarding acceleration, several studies have shown that individuals who exhibit an early onset age tend to commit crimes at a much higher rate than those with a later age of onset'' \cite{piquero2003criminal, loeber1990rate}, motivating our Effort-aware-fairness metric to consider acceleration. More broadly, our consideration of individual-level trajectory is also grounded in the group-based trajectory modeling approach in criminology, an attempt to group individuals based on their criminal trajectories over time \cite{nagin1993age, sampson2017life}, which, in our case, can be interpreted as grouping individuals based on desistance Efforts.
Our inertia term (based on childhood poverty, which is outside one's control) was found to be associated with a significantly higher ``hazard rate of being convicted of violent criminality'' \cite{sariaslan2014childhood}.

More fundamentally, we acknowledge an inherent limitation in using a concept rooted in the natural sciences (force) to characterize a social phenomenon (human Effort). Prior AI fairness literature also shows little of such intellectual borrowing. As much as we believe that multidimensional social phenomena can never have a straightforward quantification, we also believe in the power of analogical reasoning, and in this particular case, our work's analogical transfer has great potential to be fruitful.
Though we are the first to operationalize Effort as force in the context of algorithmic fairness, we are not the first to lean on the powers of analogical reasoning to facilitate social science theory development, e.g., ``Bytes provide a basis for grounding the study of social relationship and institutions in the natural sciences because their transmission or storage requires changes in energy or physical states'' \cite{turnbull2002grounding}.
We do not claim to provide an exact translation of Effort into force; instead, we aim for this characterization to be a productive metaphor that can facilitate new debate in the community on how to measure Effort in the AI fairness context better.

\section{Appendix D: Details on Predictive Models}
For CLUE predictive models, we build predictive features using three years of history (which spans data from $2013$ to $2018$), including current case details, pending cases, past arrests and convictions across degrees (e.g., misdemeanor, felony), categories (e.g., DUI, drug, public order, traffic criminal, violent), and in specific time frames (e.g., past $30$ days, ..., $1$ year, $4$ years). We set the high recidivism risk threshold to be the $62^{th}$ percentile of predicted risk scores across the felony offenders of the $25,000$ defendants in our cleaned CLUE version. This choice is inspired by the finding from the ``Pretrial Release of Felony Defendants in State Courts'' report by the DOJ (2007)\footnote{\url{https://bjs.ojp.gov/content/pub/pdf/prfdsc.pdf}} that $62\%$ of felony defendants between $1990-2004$ were released before trial, which might imply that roughly $38\%$ of felony defendants in that period were considered high risk. The ``high risk'' threshold we get with this method is approximately $0.40$.

For the SHED outcome variable, the respondent's frequency of unpaid credit card balance is recorded as the answer to the question in the survey: ``In the past 12 months, how frequently have you carried an unpaid balance on one or more of your credit cards?'' The responses are encoded as values from $\{0, 1, 2, 3\}$ where 0 indicates ``Never carried an unpaid balance (always pay in full),'' $1$ represents ``Once,'' $2$ corresponds to ``Some of the time,'' and 3 signifies ``Most or all of the time.'' This feature is then converted into a binary variable, where $0$ represents ``Never carried an unpaid balance (always pay in full)'' and $1$ indicates ``Miss credit card balance payment at least once.''

The features considered include income data from $2019$ to $2022$, employment status (e.g., employee, self-employed, full-time, part-time), household composition (e.g., the number of individuals living with the household head), demographics (e.g., education level, age), and financial status (e.g., whether the household has a checking, savings, or money market account). We split the data into a training set ($70\%$) and a testing set ($30\%$). Based on the evaluation metrics, Logistic Regression outperforms XGBoost and Random Forest in predicting missed payments. Logistic regression achieves an accuracy of 72.46\% with a weighted F1 score of 0.71. In comparison, the best-performing XGBoost model—tuned via randomized search—achieves an accuracy of 64.25\% and a weighted F1 score of 0.64.
In contrast, the random forest model achieves an accuracy of 65.70\% with a weighted F1 score of 0.64. To obtain out-of-sample risk scores for all households, we use 5-fold cross-validation. This validation method involves dividing the dataset into $5$ disjoint subsets. In each iteration, one subset is treated as test data, while the remaining $4$ subsets are used to train the model and generate predictions for the test subset. This process is repeated $5$ times, ensuring predictions are obtained for all subsets.

\section{Appendix E: Human Subjects Experiment Interfaces and Incentives}
\label{appendix: experimental-interface}
\begin{figure}[htbp]
    \centering
    \includegraphics[width=0.8\textwidth]{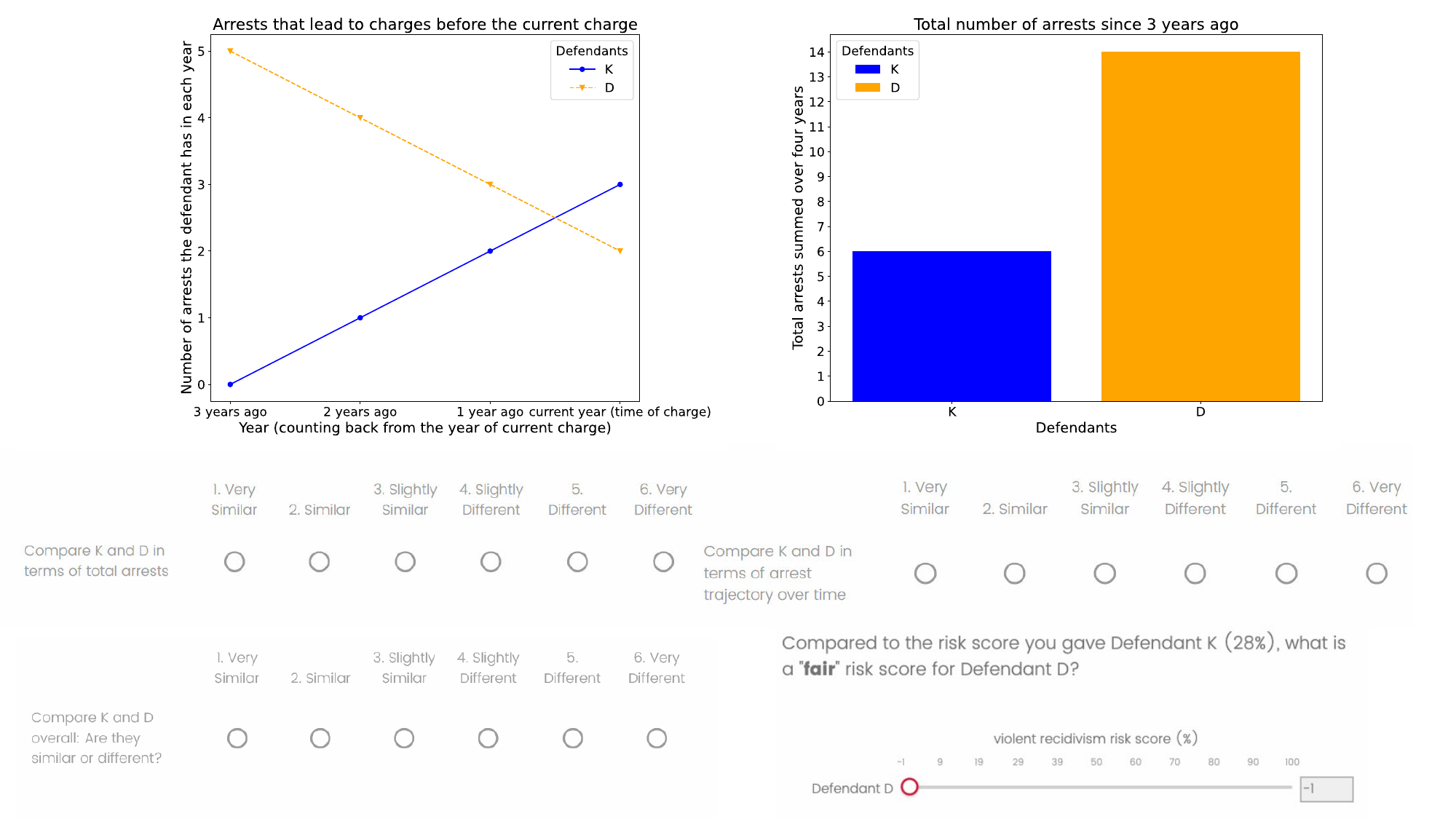}
    \caption{Example human subjects experiment interfaces to measure: (1) trajectory distance (upper left); (2) aggregate distance (upper right); (3) overall input-space distance (lower left); (4) output-space distance (lower right). Pair: ($K_1, D_1$). Condition: CLUE}
    \label{fig:CLUE_interface}
\end{figure}

\begin{figure}[htbp]
    \centering
    \includegraphics[width=0.8\textwidth]{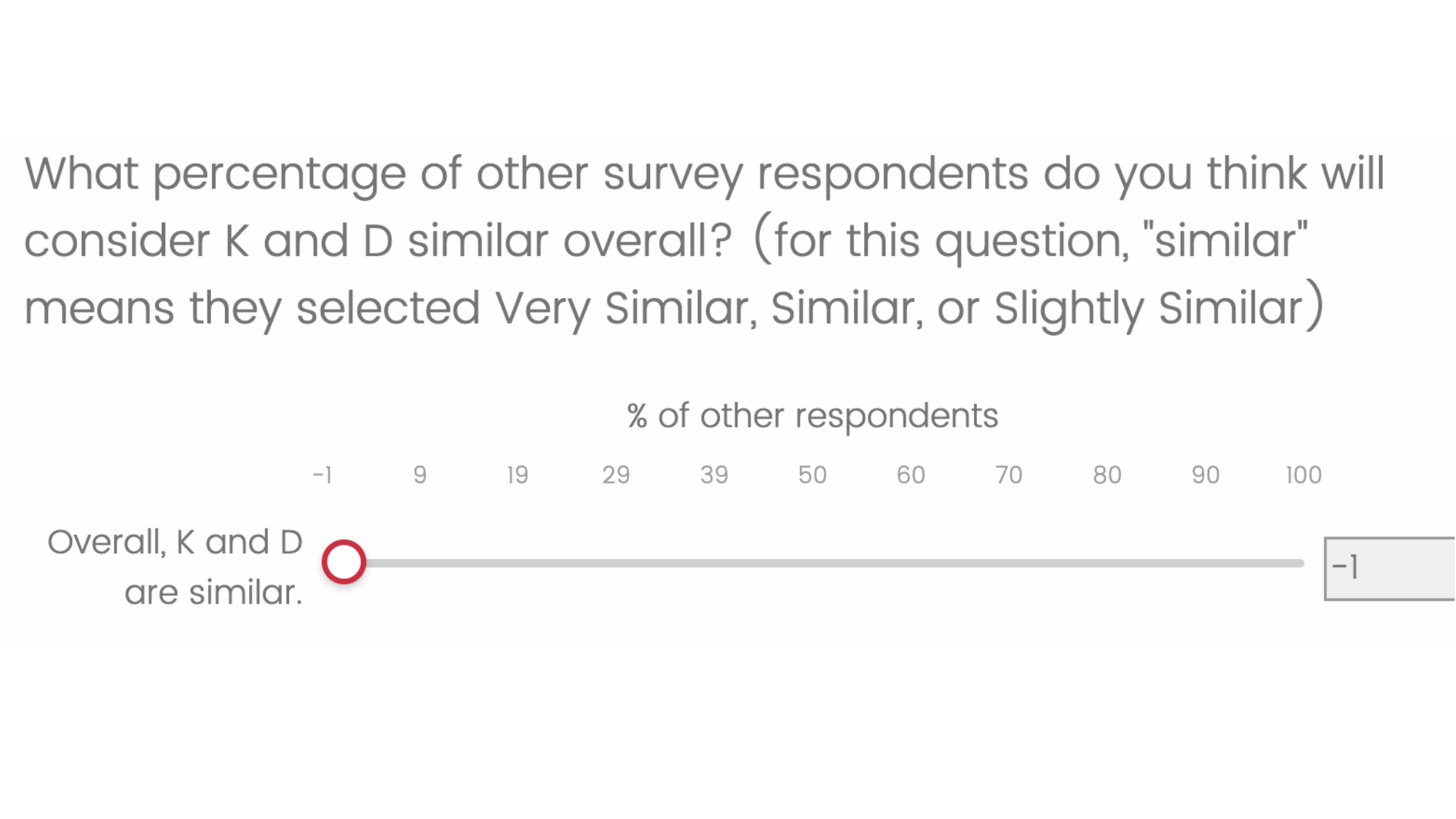}
    \caption{Example BTS frequency prediction question, following every overall input-space distance MCQ. Since there are two binarized answers (``similar'' vs. ``different'') from the 6-Likert scale, predicting the percentage of one binarized answer already implies the other.}
    \label{fig:BTS_interface}
\end{figure}

To incentivize participants to give a truthful answer to a multiple-choice question (MCQ), this method adds a follow-up question, asking each participant to predict the frequency of every answer choice in the main MCQ as selected by other participants. For each pair of MCQ and follow-up prediction question, each participant gets a BTS score = information score + prediction score. The first term, information score, will be high if the participant chooses a ``surprisingly common'' answer in the MCQ, i.e., the predicted frequency of the participant's answer (averaged across all participants' predictions) is higher than that answer's actual frequency. The intuition behind this information score is that people tend to overestimate the frequency of an answer if they think the answer is the truth. The second term, prediction score, rewards each participant's prediction if it is accurate, i.e., close to the actual frequency. The BTS method often informs participants, without revealing the math to them, that they will receive bonuses if they provide the most ``truthful'' answers, which, behind the scenes, means if they get the highest BTS scores across their answers. We give an example BTS question in Figure \ref{fig:BTS_interface}. 

As shown in Figure \ref{fig:BTS_interface}, the $6$-Likert answer choices are binarized in the BTS prediction question to reduce cognitive load.

\clearpage

\section{Appendix F: Supporting Figures and Tables}
\begin{table}[htbp]
\centering
\caption{Correlations and Significance Levels for CLUE and SHED Conditions}
\label{tab:correlation_table}
\begin{threeparttable}
\begin{tabular}{@{}lrclcc@{}}
\toprule
 &  &  &  & \textbf{CLUE Condition} & \textbf{SHED Condition} \\ 
\midrule
\multicolumn{4}{c}{\textbf{Correlation Coefficients}} & \textbf{Value ($r$)} & \textbf{Value ($r$)}  \\ 
\midrule
$r_{\text{aggr},\text{input}}$ & Aggregate & vs. & Overall Input & $0.5022^{***}$  & $0.6547^{***}$  \\ 
$r_{\text{traj},\text{input}}$ & Trajectory & vs.  & Overall Input & $0.7056^{***}$ & $0.7881^{***}$ \\
$r_{\text{aggr},\text{traj}}$ & Aggregate  & vs. & Trajectory & $0.1670^{***}$  & $0.4277^{***}$  \\
$r_{\text{aggr},\text{output}}$ & Aggregate  & vs.  & Output & $0.2089^{***}$  & $0.2507^{***}$  \\
$r_{\text{traj},\text{output}}$ & Trajectory   & vs.  & Output & $0.4218^{***}$  & $0.4986^{***}$  \\
\bottomrule
\end{tabular}
\begin{tablenotes}
\item \textbf{Note:} Significance levels: *** \( p < 0.001 \), ** \( p < 0.01 \), * \( p < 0.05 \). All correlation coefficients remain significant after applying the Bonferroni correction. 
\end{tablenotes}
\end{threeparttable}
\end{table}

\begin{table}[htbp]
\centering
\caption{Hotelling's t-Test and Steiger's Z-test Results for CLUE and SHED Conditions}
\label{tab:test_results}
\begin{threeparttable}
\begin{tabular}{@{}lrclcc@{}}
\toprule
& & & & \textbf{CLUE Condition} & \textbf{SHED Condition} \\ 
\midrule
\textbf{Test} & \multicolumn{3}{c}{\textbf{Comparison}} & \textbf{Statistic}  & \textbf{Statistic}  \\ 
\midrule
Hotelling's t-test & $r_{\text{traj},\text{input}}$ & vs. & $r_{\text{aggr},\text{input}}$ & $7.9758^{***}$ & $7.3360^{***}$ \\
Hotelling's t-test & $r_{\text{traj},\text{output}}$ & vs. & $r_{\text{aggr},\text{output}}$ & $5.5161^{***}$ & $7.9612^{***}$ \\
Steiger's Z-test & $r_{\text{traj},\text{input}}$ & vs. & $r_{\text{aggr},\text{input}}$ & $6.9077^{***}$ & $6.5541^{***}$ \\
Steiger's Z-test  & $r_{\text{traj},\text{output}}$ & vs. & $r_{\text{aggr},\text{output}}$ & $5.3461^{***}$ & $7.6466^{***}$  \\
\bottomrule
\end{tabular}
\begin{tablenotes}
\item \textbf{Note:} Significance levels: *** \( p < 0.001 \), ** \( p < 0.01 \), * \( p < 0.05 \). All statistics remain significant after applying the Bonferroni correction.
\end{tablenotes}
\end{threeparttable}
\end{table}

\begin{table}[htbp]
\centering
\caption{95\% Confidence Intervals for CLUE and SHED Conditions}
\label{tab:ci_table}
\begin{threeparttable}
\begin{tabular}{@{}lrclcc@{}}
\toprule
&  &  & & \textbf{CLUE Condition} & \textbf{SHED Condition}\\ 
\midrule
\multicolumn{4}{c}{\textbf{Differences in Correlation Coefficients}} & \textbf{Confidence Interval} & \textbf{Confidence Interval} \\ 
\midrule
\multicolumn{4}{l}{$\Delta r_{\text{input}}$=$r_{\text{aggr},\text{input}} - r_{\text{traj},\text{input}}$} & $[-0.2780, -0.1280]$ & $[-0.1950, -0.0730]$ \\
\multicolumn{4}{l}{$\Delta r_{\text{output}}$=$r_{\text{aggr},\text{output}} - r_{\text{traj},\text{output}}$} & $[-0.3000, -0.1280]$ & $[-0.3330, -0.1610]$\\
\bottomrule
\end{tabular}
\end{threeparttable}
\end{table}

\begin{table}[htbp]
\setlength{\abovecaptionskip}{6pt} 
\setlength{\belowcaptionskip}{6pt} 
\centering
\caption{Regression Results for CLUE and SHED Models}
\label{tab:regression_results}
\begin{threeparttable}
\setlength{\tabcolsep}{2pt} 
\renewcommand{\arraystretch}{0.8} 
\small 
\begin{tabular}{@{}l@{}cccccc@{}}
\toprule
& \multicolumn{2}{c}{\textbf{CLUE Condition}} & \multicolumn{2}{c}{\textbf{SHED Condition}} & \multicolumn{2}{c}{\textbf{Moderation Effects}} \\
& $d$ & $D$ & $d$ & $D$ & $d$ & $D$ \\ 
\midrule
Intercept      & 0.4058***  & 0.0156     & 0.3224*** & 0.0060     & 0.3224***  & 0.0060      \\
                        & (0.0833)   & (0.0164)   & (0.0698)  & (0.0162)   & (0.0745)   & (0.0159)    \\ 
$d_S$: Aggregate    & 0.3580***  & 0.0070**   & 0.3182*** & -0.0132*** & 0.3182***  & -0.0132***  \\
                        & (0.0179)   & (0.0035)   & (0.0154)  & (0.0036)   & (0.0164)   & (0.0035)    \\ 
$d_E$: Trajectory    & 0.5454***  & 0.0403***  & 0.6114*** & 0.0565***  & 0.6114***  & 0.0565***   \\
                        & (0.0171)   & (0.0034)   & (0.0169)  & (0.0039)   & (0.0181)   & (0.0039)    \\ 
$C_{\text{CLUE}}$          &            &            &           &            & 0.0834     & 0.0096      \\
                        &            &            &           &            & (0.1084)   & (0.0231)    \\ 
$d_S \times C_{\text{CLUE}}$   &            &            &           &            & 0.0399*    & 0.0201***   \\
                        &            &            &           &            & (0.0236)   & (0.0050)    \\ 
$d_E \times C_{\text{CLUE}}$   &            &            &           &            & -0.0661*** & -0.0163***  \\
                        &            &            &           &            & (0.0242)   & (0.0052)    \\ 
\midrule
\textbf{R-squared}      & 0.6580     & 0.1534     & 0.7794    & 0.1975     & 0.7266     & 0.1769      \\ 
\textbf{Adj. R-squared} & 0.6573     & 0.1515     & 0.7789    & 0.1957     & 0.7258     & 0.1746      \\ 
\textbf{F-statistic}    & 863.05     & 81.27      & 1563.19   & 108.88     & 947.11     & 76.62       \\ 
\textbf{Obs.}              & 900        & 900         & 888        & 888         & 1788             & 1788               \\
\bottomrule
\end{tabular}
\begin{tablenotes}
\item[] \textbf{Notes:} Standard errors in parentheses. * \(p < 0.1\), ** \(p < 0.05\), *** \(p < 0.01\).
\end{tablenotes}
\end{threeparttable}
\end{table}

\begin{table}[htbp]
 \centering
\caption{Traditional (Effort-UNaware) Demographic Parity on SHED in terms of mean predicted risk. A demographic group must have at least 10 individuals to be considered.}
 \label{tab:GF_SHED}
\begin{tabular}{lccc}
\toprule
\textbf{Model}            & \textbf{Race parity} & \textbf{Sex parity} & \textbf{Age parity} \\
\midrule
XGBoost             & 0.48                            & 0.99                           & 0.78                           \\ 
Logistic regression & 0.49                            & 0.94                           & 0.78                           \\ 
Random forest       & 0.57                            & 0.99                           & 0.83    \\
\bottomrule
\end{tabular}
\end{table}

\clearpage
\section{Appendix G: Demonstrating Effort-aware Fairness metrics on CLUE dataset}
We present EaGF and EaIF metrics computed for the CLUE dataset, analogous to those for SHED in the main text. Regarding EaIF, most steps (pairwise fairness score and average EaIF score per model) remain similar to those in the SHED computation, except that we slightly modify the scaling factor $\lambda$ in the aggregate (summing) function $S$ applied on the past arrests feature (or the most recent cumulative arrests) $X_3=\sum_{t=0}^{3} x_t$ as follows:
\begin{align}
    S(\mathbf{x}) 
    = 2  \cdot \sigma\left(\frac{\sum_{t=0}^{3} x_t}{1 \text{ arrest}}\right)-1 
    = 2  \cdot \sigma\left(\frac{X_3}{\lambda}\right)-1
\end{align}
Table \ref{tab:EaIF_CLUE} shows that Light GBM and Logistic regression are the most and least fair according to our EaIF metric, respectively.

\begin{table}[htbp]
 \centering
 \caption{Effort-aware Individual Fairness Results on CLUE (equal weights are $0.5$ and $0.5$; human study weights are regression coefficients to measure impacts on overall input-space distance, normalized to $0.6037$ for Effort and $0.3963$ for aggregate)}
 \label{tab:EaIF_CLUE}
\begin{tabular}{lcc}
\toprule
           & \multicolumn{2}{c}{\textbf{EaIF}} \\
\textbf{Model}                  & Equal weights    & Human study weights  \\
\midrule
Light GBM             & 0.951                             & 0.951                                    \\ 
Logistic regression & 0.928                            & 0.929                                     \\
Decision tree      & 0.946                            & 0.946 \\ 
\bottomrule
\end{tabular}
\end{table}

\begin{figure}[htbp]
    \centering
    \begin{subfigure}[b]{0.5\textwidth}
        \centering
        \includegraphics[width=\textwidth]{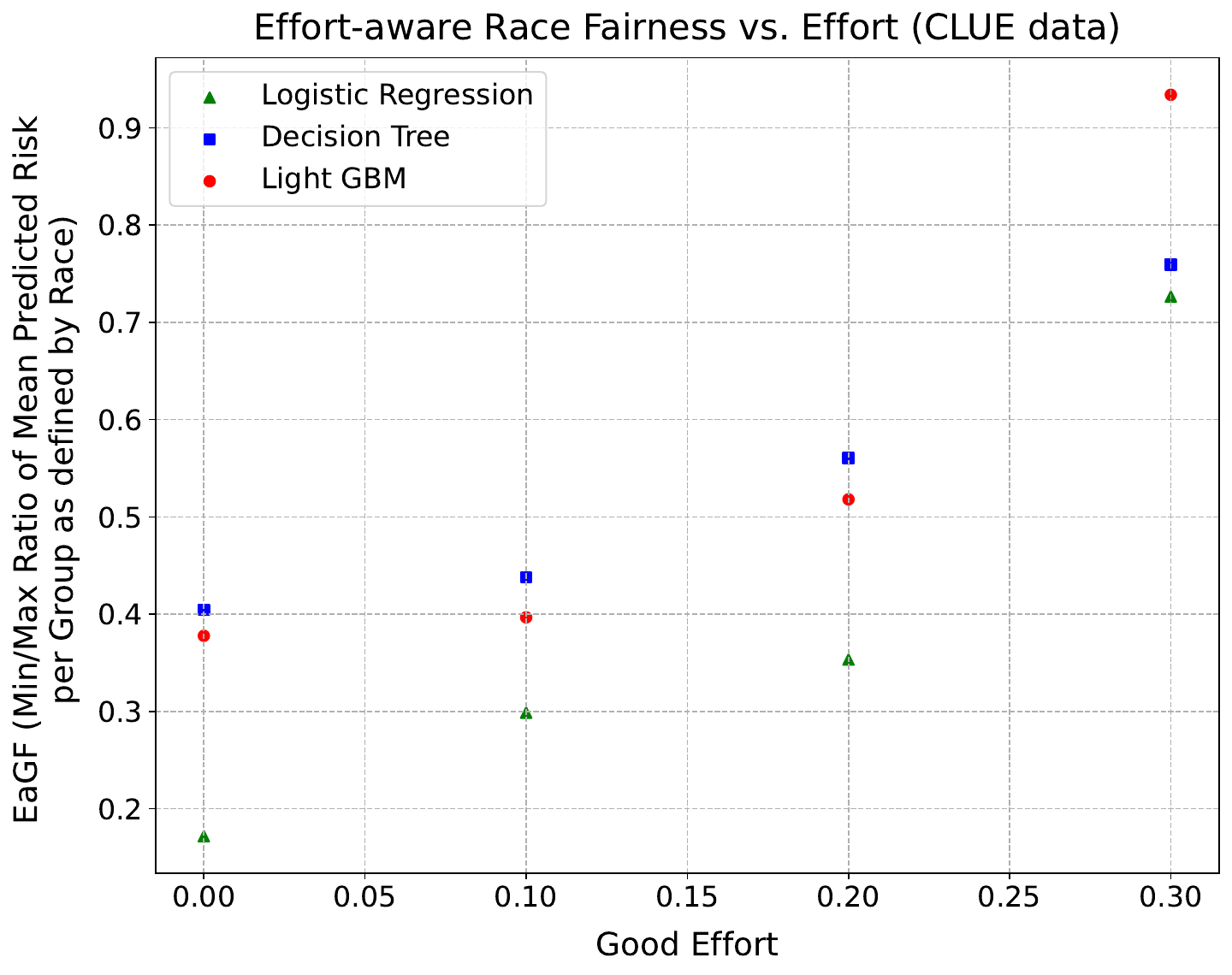}
        \caption{Race parity}
        \label{subfig:CLUE_EaGF_App1_Race}
    \end{subfigure}
    \hfill
    \begin{subfigure}[b]{0.5\textwidth}
        \centering
        \includegraphics[width=\textwidth]{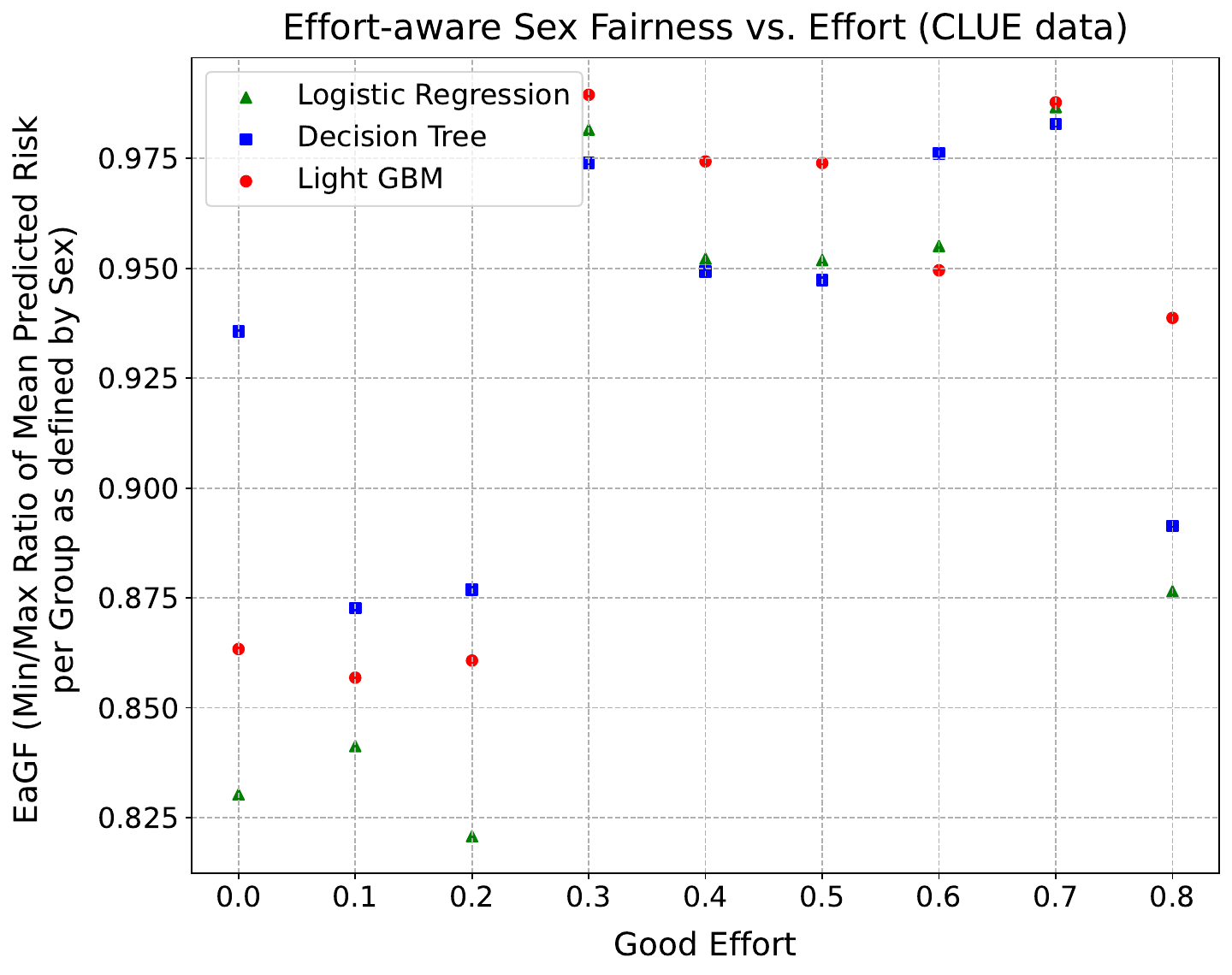}
        \caption{Sex parity}
        \label{subfig:CLUE_EaGF_App1_Sex}
    \end{subfigure}
    \hfill
    \begin{subfigure}[b]{0.5\textwidth}
        \centering
        \includegraphics[width=\textwidth]{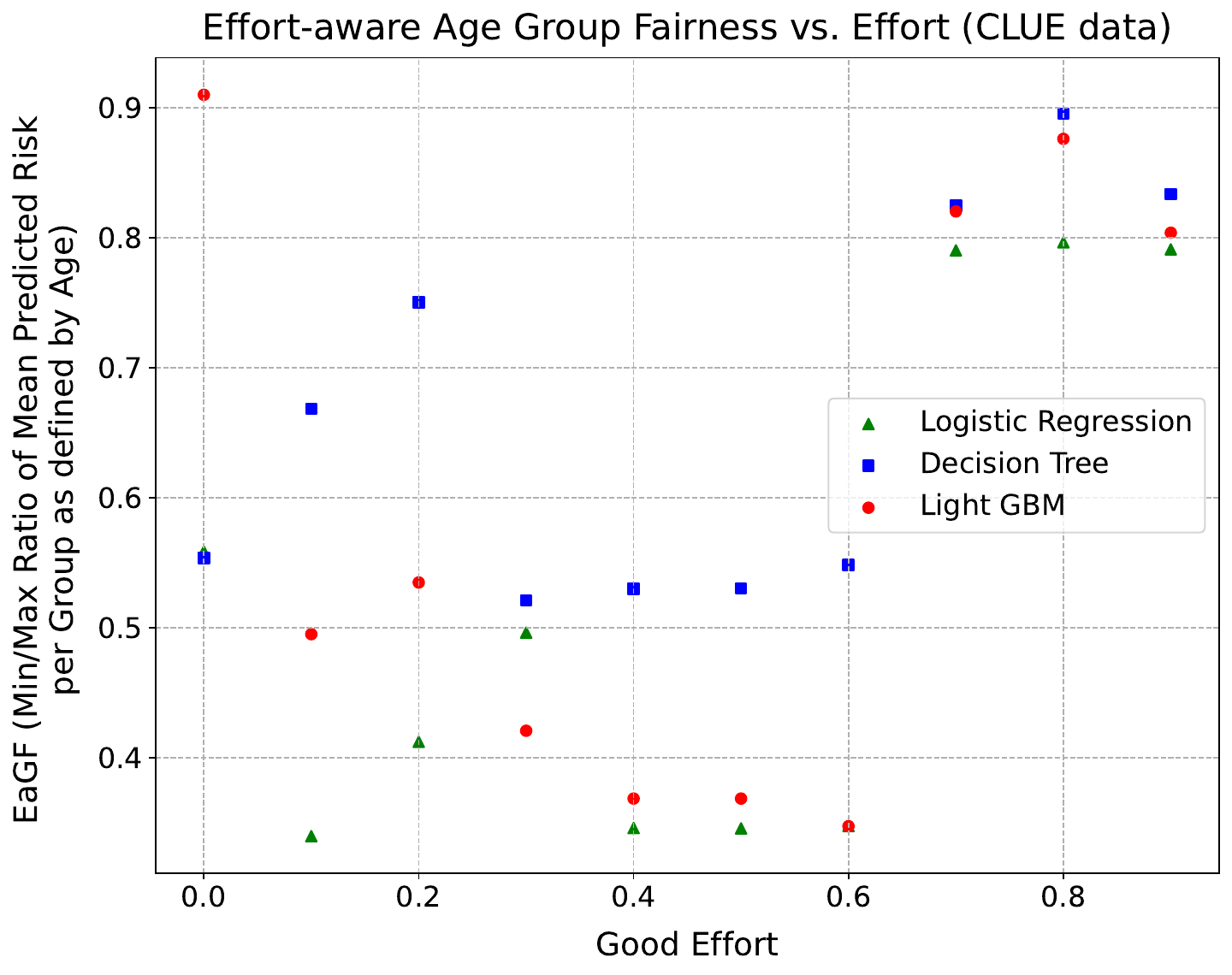}
        \caption{Age group parity}
        \label{subfig:CLUE_EaGF_App1_Age}
    \end{subfigure}
    \caption{Effort-aware Group Parity (in terms of mean predicted risk) as a function of Effort (CLUE data). Bin length is 0.1. Only demographic groups with at least 10 data points are considered. The subplots' x-axes might differ since not all Effort bins within the 0-1 range have enough data points to compute EaGF. Decision tree achieves the best age group parity (compared to logistic regression and light GBM) in most Effort bins. Higher-Effort bins get better race parity across all three AI models.}
    \label{fig:CLUE_EaGF_App1}
\end{figure}

Regarding EaGF, although Table \ref{tab:GF_CLUE} on traditional, Effort-unaware demographic parity may give an impression that decision tree achieves the best group fairness across all three demographic features (race, sex, and age group), more nuanced investigation of EaGF in Figure \ref{fig:CLUE_EaGF_App1} shows that decision tree only achieves consistently the best EaGF results among age groups. Between sexes, while decision tree achieves the best EaGF at the lower-Effort bins ($0$ to $0.3$), Light GBM achieves the best EaGF at most of the higher-Effort bins ($0.3$ to $0.9$), indicating that Light GBM might be a better fit if the application prioritizes equality between males and females of similar Efforts while incentivizing both sexes to exert more future Efforts. With respect to race, all three models seem to give the right incentive, where higher-effort bins generally suffer less demographic disparity. 

\begin{table}[htbp]
 \centering
 \caption{Traditional (Effort-UNaware) Demographic Parity on CLUE in terms of mean predicted risk. A demographic group must have at least 10 individuals to be considered.}
 \label{tab:GF_CLUE}
\begin{tabular}{lccc}
\toprule
\textbf{Model}            & \textbf{Race parity} & \textbf{Sex parity} & \textbf{Age parity} \\
\midrule
Light GBM           & 0.44                            & 0.91                           & 0.45                           \\ 
Logistic regression & 0.30                            & 0.88                           & 0.44                           \\ 
Decision tree       & 0.49                            & 0.92                           & 0.49    \\

\bottomrule
\end{tabular}
\end{table}

\end{document}